\documentclass[acmsmall,nonacm]{acmart}

\usepackage{algorithm}
\usepackage{algpseudocode}
\usepackage{multirow}
\usepackage{placeins}

\graphicspath{{fig/}}
\AtBeginDocument{}

\setcopyright{none}
\begin{document}
\title{Efficient Remote Sensing Image-Text Retrieval via Interaction-Free Recall and Text-Guided Reranking}
\renewcommand{\shorttitle}{FTF for Efficient Remote Sensing Image--Text Retrieval}

\author{Xi Chen}
\email{chenxi00233@whu.edu.cn}
\affiliation{\institution{School of Computer Science, Wuhan University}\city{Wuhan}\country{China}}

\author{Xu Chen}
\authornote{Corresponding author.}
\email{xuchen@whu.edu.cn}
\affiliation{\institution{School of Computer Science, Wuhan University}\city{Wuhan}\country{China}}

\author{Xiangyang Jia}
\email{jxy@whu.edu.cn}
\affiliation{\institution{School of Computer Science, Wuhan University}\city{Wuhan}\country{China}}

\author{Wei Wang}
\email{wangwei@bigai.ai}
\affiliation{\institution{Beijing Institute for General Artificial Intelligence (BIGAI)}\city{Beijing}\country{China}}

\author{Xu Zhang}
\email{zhangx0802@whu.edu.cn}
\affiliation{\institution{School of Computer Science, Wuhan University}\city{Wuhan}\country{China}}

\author{Zhenyuan Sun}
\email{zhenyuansun@whu.edu.cn}
\affiliation{\institution{School of Computer Science, Wuhan University}\city{Wuhan}\country{China}}

\renewcommand{\shortauthors}{Chen et al.}

\begin{abstract}
Remote sensing (RS) image--text retrieval plays a critical role in understanding massive RS imagery. However, the dense multi-object distribution and complex backgrounds in RS imagery make it difficult to simultaneously achieve fine-grained cross-modal alignment and efficient retrieval.
  Existing methods either rely on complex cross-modal interactions that lead to low retrieval efficiency, or depend on large-scale vision--language adaptation, which requires massive data and computational resources.
  To address these issues, we develop a fast-then-fine (FTF) retrieval framework centered on interaction-free multi-granular recall and parameter-free text-guided cross-modal reranking with the Balanced Text-guided Interaction Block (BTIB).
  Specifically, interaction-free multi-granular recall pre-computes gallery embeddings and matches them with the corresponding query embedding without full-gallery token-level text--image interaction, while text-guided cross-modal reranking with BTIB refines local semantic correspondence within the top-$K$ candidate set.
  Furthermore, an inter- and intra-modal loss is designed to jointly optimize cross-modal alignment across multi-granular representations.
  Extensive experiments on public benchmarks demonstrate that the FTF achieves competitive retrieval accuracy while significantly improving retrieval efficiency compared with existing methods.
\end{abstract}



\maketitle

\section{Introduction}
Remote sensing (RS) imagery has grown rapidly with the continuous development of earth observation platforms, generating massive volumes of visual data~\cite{Astruc_2025_CVPR}.  Efficient access to and understanding of RS imagery have become increasingly important~\cite{li2025lhrs}. 
Image--text retrieval provides an effective way to retrieve RS images through textual queries by linking visual content with semantic descriptions~\cite{23acmmmPIR,23tgrsKAMCL,25tgrsCLGSA,24acmmmEBAKER}. 
However, RS imagery typically exhibits high spatial resolution and complex background composition, where diverse objects and fine structural details coexist within a single scene, making it difficult to simultaneously achieve efficient retrieval and fine-grained alignment.

\begin{figure}[!htb]
	\centering
	\includegraphics[width=\linewidth,height=0.23\textheight,keepaspectratio]{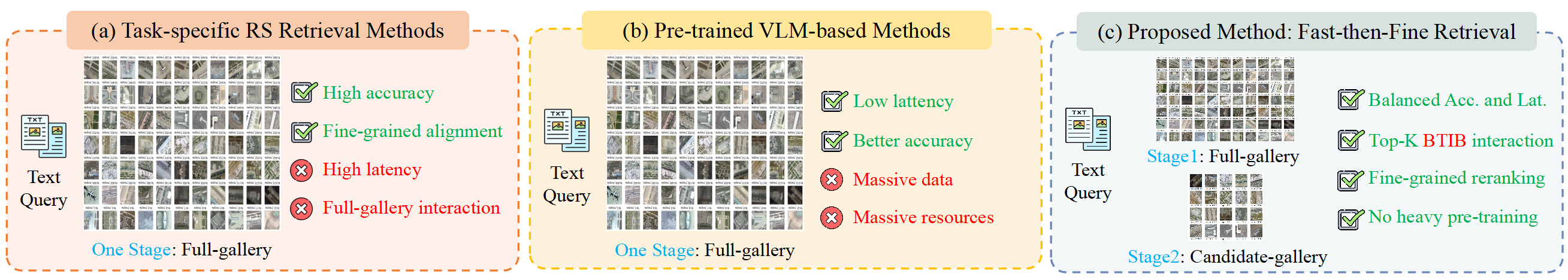}
	\caption{
	Comparison of three paradigms for RS image--text retrieval.
	(a) Task-specific methods perform fine-grained matching via complex full-gallery token-level text--image interaction, but suffer from high latency.
	(b) VLM-based methods achieve strong generalization, but often require massive data and computational resources.
	(c) Our FTF performs interaction-free multi-granular recall followed by text-guided cross-modal reranking with BTIB, achieving a better balance between retrieval accuracy and efficiency.
	}
	\Description{Three retrieval paradigms are compared: full-gallery task-specific interaction, large vision--language model adaptation, and the proposed two-stage FTF framework.}
	\label{fig:motivation}
\end{figure}

Recent years have witnessed substantial progress in RS image--text retrieval~\cite{25tmmCPPMN,25acmmmEAPA}. 
On one hand, as illustrated in Fig.~\ref{fig:motivation}(a), many methods leverage complex cross-modal interactions to achieve fine-grained alignment between RS images and text, which improves retrieval accuracy but often introduces significant computational overhead and limits retrieval efficiency~\cite{22tgrsGaLR,23sensorsFAAMI,24tgrsMSCMAT,24tgrsIERR}. 
On the other hand, as illustrated in Fig.~\ref{fig:motivation}(b), recent VLM-based methods further improve retrieval performance by leveraging strong pretrained vision--language representation models, but these methods typically require massive data and computational resources~\cite{23tgrsPETL,24aaaiskysrcipt,24iclrGRAFT,24tgrsGeoRSCLIP}. More recently, MLLM-assisted methods such as Caption2CLIP~\cite{26tmmCaption2CLIP} and GraVLM~\cite{26tgrsGraVLM} further improve retrieval accuracy through caption enrichment or semantic entity extraction. However, these additional MLLM-based steps may introduce considerable inference overhead in practical deployment.
As a result, achieving a practical balance between accuracy and efficiency remains a critical challenge for RS image--text retrieval. 

To address the above challenges, we propose a fast-then-fine (FTF) two-stage retrieval framework for RS image--text retrieval, as shown in Fig.~\ref{fig:motivation}(c). 
Considering the high resolution, dense objects, and fine-grained spatial descriptions in RS image--text retrieval, applying token-level text--image interaction to the full gallery is often unnecessarily expensive. FTF therefore separates interaction-free multi-granular recall from text-guided cross-modal reranking with BTIB: the recall stage still uses the query text embedding for similarity matching, but avoids full-gallery token-level text--image interaction. Unlike one-stage local-global interaction models that apply expensive cross-modal reasoning throughout retrieval, FTF explicitly controls where the costly interaction is performed, placing it only after the top-$K$ candidate set has been obtained. This design reduces expensive full-gallery token-level text--image interaction while preserving the ability to achieve fine-grained cross-modal alignment. 
Compared with existing task-specific methods that mainly improve retrieval accuracy through increasingly complex matching, the proposed framework offers a practical way to improve the accuracy--efficiency balance in RS image--text retrieval~\cite{22tgrsGaLR,23acmmmPIR,24tgrsVGSGN,25tgrsSIPN,25tgrsFSSN}. Its focus is not merely to introduce a cascaded retrieval pipeline, but to coordinate scalable multi-granular candidate recall with lightweight, query-conditioned refinement under a controllable candidate budget.

The main contributions of this work are summarized as follows:
\begin{itemize}
	\item 
	We propose an efficiency-oriented FTF framework that coordinates interaction-free multi-granular recall and candidate-level fine-grained refinement. By constructing pre-computable gallery embeddings from complementary scene and local semantics and restricting text--image interaction to the top-$K$ candidates, FTF provides an explicit and adjustable accuracy--efficiency trade-off.
	
	\item 
	We design the Balanced Text-guided Interaction Block (BTIB), a parameter-free reranking module that combines text-conditioned visual normalization with visual-token balancing normalization. This complementary normalization refines candidate-level semantic correspondence while avoiding additional learnable parameters.
	
	\item
	We introduce an Inter- and Intra-Modal (I$^2$M) Loss that jointly optimizes recall and reranking representations by coupling cross-modal alignment with intra-modal neighborhood consistency across multiple visual granularities.
	
	\item 
	We conduct extensive experiments on RSITMD, RSICD, and UCM, including candidate coverage analysis, reranking alternative comparison, and accuracy--efficiency evaluation, demonstrating that FTF achieves competitive retrieval accuracy with substantially lower inference cost.
\end{itemize}

\section{Related Work}
\subsection{Task-Specific RS Retrieval Methods} Existing RS image--text retrieval methods mainly improve retrieval accuracy by designing increasingly complex cross-modal interaction mechanisms, thereby enhancing semantic alignment between RS imagery and textual descriptions. Representative methods, including CAMP~\cite{2019iccvCAMP}, LW-MCR~\cite{21tgrsLW-MCR}, AMFMN~\cite{22tgrsAMFMN}, KAMCL~\cite{23tgrsKAMCL}, and SCAN~\cite{2018eccvSCAN}, focus on establishing semantic correspondence through cross-modal projection, attention-based matching~\cite{25patrecCMA}, semantic consistency learning~\cite{26tmmCIGNN}, and semantic correlation modeling~\cite{25nnUSSL,25airCSAN}, which lay an important foundation for RS image--text retrieval.
Building on this line, more recent methods, such as GaLR~\cite{22tgrsGaLR}, PIR~\cite{23acmmmPIR}, FAAMI~\cite{23sensorsFAAMI}, MTGFE~\cite{23RemoteSensingMTGFE}, MSCMAT~\cite{24tgrsMSCMAT}, IERR~\cite{24tgrsIERR}, VGSGN~\cite{24tgrsVGSGN}, SIPN~\cite{25tgrsSIPN}, and FSSN~\cite{25tgrsFSSN}, further strengthen fine-grained cross-modal alignment through local-to-global matching~\cite{25tgrsCLGSA,25cvprGOAL}, relation-aware modeling, multi-granular feature learning, and richer semantic interaction~\cite{23sigirMGRL,25neunetSFE,24asocCCCL}. 
Recent studies have also investigated RS image--text retrieval directly~\cite{26arxivDARAD}. SSJDN jointly decouples scale-specific and semantic features to improve cross-modal representation, whereas EEDR addresses incorrectly correlated and weakly correlated image--text pairs through correspondence rectification and diversity-aware feature decoupling~\cite{23tommSSJDN,25tommEEDR}.
These methods indicate that fine-grained modeling is crucial for handling the high spatial resolution and complex background characteristics of RS imagery~\cite{Asokan_2025_CVPR,ma2025multimodal}. However, these methods apply complex cross-modal interaction directly over the full candidate space, which limits retrieval efficiency in practical retrieval scenarios.

\subsection{VLM-Based Methods} In addition to task-specific RS image--text retrieval methods, recent studies have further improved retrieval performance by leveraging strong vision--language representations~\cite{21icmlCLIPBASE,22icmlBLIP,23icmlBLIP2}. 
Representative methods, including CLIP-Base~\cite{21icmlCLIPBASE} and PETL~\cite{23tgrsPETL}, transfer general-purpose vision--language representations to the RS domain, while SkyScript~\cite{24aaaiskysrcipt}, GRAFT~\cite{24iclrGRAFT}, RemoteCLIP~\cite{24tgrsremoteclip}, and GeoRSCLIP~\cite{24tgrsGeoRSCLIP} further enhance retrieval accuracy through RS-oriented adaptation. 
These methods demonstrate that strong vision--language representations can improve cross-modal representation learning and semantic alignment~\cite{25cvprPostAlign,24eccvSILC,25tpamiGLSCL,24prModalInteraction,26tipPerceiveIR} between RS imagery and textual descriptions. 
However, their performance usually depends on massive data and computational resources~\cite{wei2025hqclip,huang2025noiserobust}. 
More recently, MLLM-assisted methods further exploit generative semantic understanding to enhance RS image--text retrieval. For example, Caption2CLIP enriches visual representations with MLLM-generated hierarchical captions and language modeling ability from LLaMA\cite{24arxivLlama3}, while GraVLM~\cite{26tgrsGraVLM} uses DeepSeek~\cite{24arxivDeepSeekVL} to extract semantic entities and construct graph-aligned cross-modal representations. Although these generative semantics are helpful for improving retrieval accuracy, they introduce additional caption generation, entity parsing, or reasoning steps.
As a result, these VLM- and MLLM-based directions do not directly address how to achieve efficient fine-grained matching without relying on large-scale data collection, high-cost model adaptation, or additional generative inference.

\subsection{Efficient Cross-Modal Retrieval}

Efficient cross-modal retrieval~\cite{26arxivMultiBit} commonly relies on lightweight dual-encoder models, where image and text embeddings are pre-computed and matched through fast similarity search~\cite{21icmlCLIPBASE,25aclOMGM}. This design enables scalable gallery search, but compact global representations may lose fine-grained matching accuracy, especially for RS image--text retrieval with complex backgrounds, multiple objects, and rich spatial relationships~\cite{22tgrsGaLR,24tgrsMSCMAT,24tgrsIERR,25tgrsFSSN}. Recent general-domain studies improve image--text matching through authentic negative repository learning and lightweight cluster refinement~\cite{25tommReAL}. These methods improve representation learning and matching robustness, but they are not designed for the RS-specific accuracy--efficiency trade-off.

To balance efficiency and accuracy, recall-then-rerank pipelines have been explored in several retrieval scenarios, where a lightweight recall stage first selects candidates and a stronger reranking model then performs more detailed semantic matching~\cite{21neuripsALBEF,22icmlBLIP}. Efficient reranking has also been reformulated as sparse graph propagation to reduce the computational cost of neighborhood-based refinement~\cite{26tommGNNRerank}. In RS retrieval, two-stage reranking has been studied for single-modal image retrieval~\cite{tsr2017}, while CFITR directly addresses RS image--text retrieval by combining dual-stream hash-based coarse retrieval with single-stream Transformer-based fine matching~\cite{25grslCFITR}. FTF follows a different design: it constructs multi-granular CLIP representations for interaction-free candidate recall and applies the parameter-free BTIB only to the top-$K$ candidates. This formulation avoids hash-based retrieval and learnable cross-modal reranking layers while retaining fine-grained, query-conditioned refinement under an explicit candidate budget.

\section{Methods}

\subsection{Problem Formulation}

Let $\mathcal{D} = \{(I_i, x_i)\}_{i=1}^{N}$ denote a training set of paired RS images $I_i$ and textual descriptions $x_i$, where $N$ is the number of samples.  
Given a textual query $x_q$, the goal of RS image--text retrieval is to return a ranked list of images from a large gallery $\mathcal{G}$ that are semantically relevant to $x_q$.  

Each textual description $x_i$ is encoded into a token-level representation $T_i$ and a global text embedding $t_i$ using a text encoder.  
For a given textual query, each candidate image $I_i$ is represented by three visual embeddings: (1) the interaction-free multi-granular recall embedding $\mathbf{v}_{1,i}$; (2) the text-guided coarse embedding $\mathbf{v}_{2,i}$ for reranking; and (3) the text-guided fine embedding $\mathbf{v}_{3,i}$ for reranking.  

The overall objective is to learn embeddings $\mathbf{v}_1, \mathbf{v}_2, \mathbf{v}_3$ and $t$ such that (i) $\mathbf{v}_1$ efficiently identifies the top-$K$ candidate set in the interaction-free multi-granular recall stage, and (ii) $\mathbf{v}_2$ and $\mathbf{v}_3$ support text-guided cross-modal reranking through complementary coarse-grained and fine-grained semantic alignment with the query.
For both text and visual encoding, we adopt transformer-based encoders. Detailed backbone settings are provided in the experimental section.

\subsection{Overview}

We denote the gallery of RS images as $\mathcal{G}$ and a textual query as $x_q$.  
The FTF framework aims to retrieve a ranked subset of candidate images $\mathcal{C} \subset \mathcal{G}$ that are semantically relevant to $x_q$ while maintaining high efficiency.  
Formally, the interaction-free multi-granular recall stage computes a visual mapping
\begin{equation}
	\mathbf{v}_1 = f_{\mathrm{recall}}(I), \quad I \in \mathcal{G},
	\label{eq:recall_mapping}
\end{equation}
to produce multi-granular embeddings for efficient candidate selection, where $f_\mathrm{recall}$ aggregates coarse and fine patch token embeddings. The interaction-free design allows gallery visual embeddings to be pre-computed without query-specific token-level text--image interaction, while candidate retrieval is performed through embedding-level similarity with the query.
For each candidate image, the text-guided cross-modal reranking stage then performs BTIB-based interaction between its visual tokens and the query textual tokens:
\begin{equation}
	\mathbf{v}_2, \mathbf{v}_3 = f_{\mathrm{BTIB}}(\mathbf{H}^c, \mathbf{H}^f, T_q),
	\label{eq:rerank_mapping}
\end{equation}
where $\mathbf{v}_2$ and $\mathbf{v}_3$ denote coarse and fine text-guided embeddings generated by BTIB.  

\begin{figure}[!htb]
	\centering
	\includegraphics[width=1.0\linewidth,height=0.30\textheight,keepaspectratio]{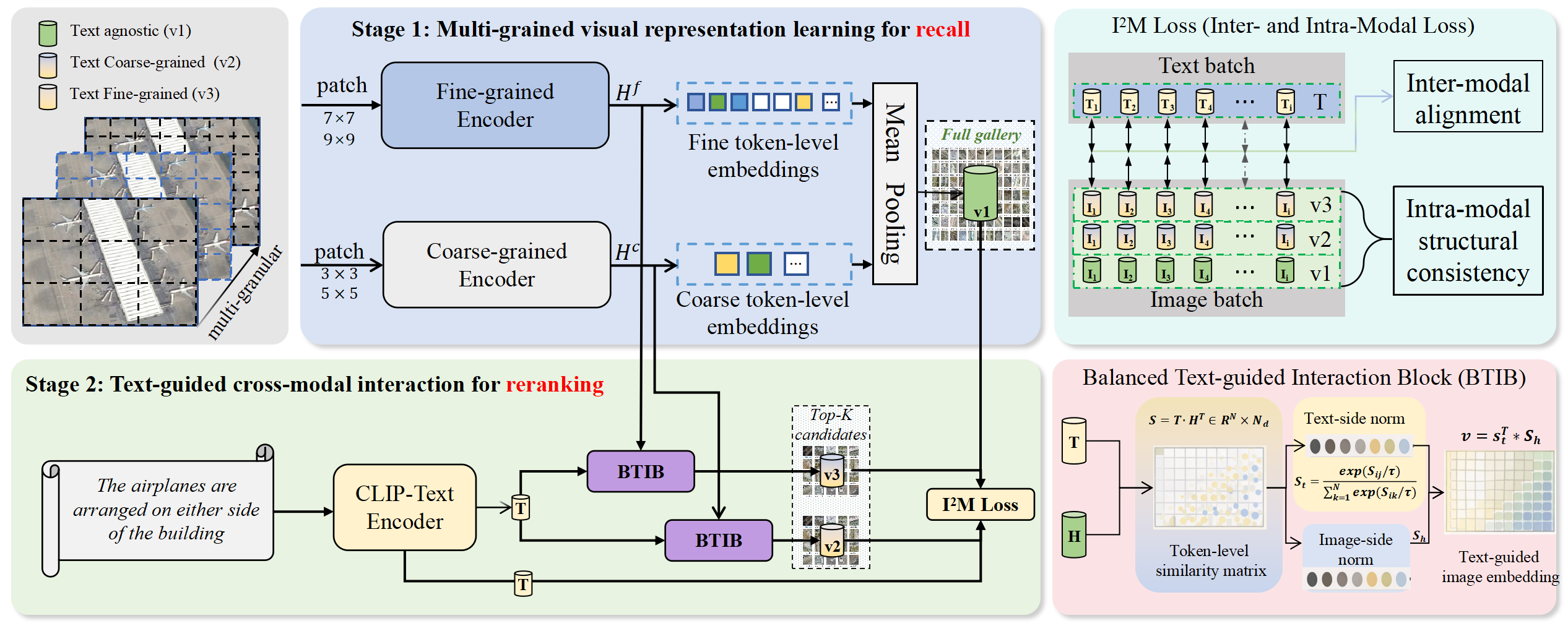}
	\caption{The overall FTF two-stage retrieval framework. 
		Multi-granular visual embeddings $\mathbf{v}_1$ are used for interaction-free multi-granular recall, where query texts are matched with pre-computed gallery features without full-gallery token-level text--image interaction,
		while BTIB generates $\mathbf{v}_2$ and $\mathbf{v}_3$ in the text-guided cross-modal reranking stage. 
		I$^2$M Loss jointly optimizes cross-modal alignment and intra-modal structural consistency.}
	\Description{The FTF pipeline contains interaction-free multi-granular recall, top-K candidate selection, BTIB-based text-guided reranking, and inter- and intra-modal training objectives.}
	\label{fig:framework}
\end{figure}

Fig.~\ref{fig:framework} illustrates the overall framework.  
Given a query from either modality, interaction-free multi-granular recall first selects the top-$K$ candidates from the opposite-modality target gallery by comparing query and target embeddings, avoiding full-gallery token-level text--image interaction.  
The text-guided cross-modal reranking stage then applies BTIB only to the selected query--candidate pairs and refines the final ranking with the text-guided coarse and fine embeddings $\mathbf{v}_2$ and $\mathbf{v}_3$.  
This two-stage design balances retrieval efficiency and retrieval accuracy, as fine-grained cross-modal interaction is restricted to the top-$K$ candidate set.

\begin{algorithm}[t]
\caption{Fast-then-Fine Retrieval with BTIB}
\label{alg:ftf}
\begin{algorithmic}[1]
	\Require Text query $x_q$, image gallery $\mathcal{G}$, candidate size $K$
	\Ensure Ranked image list
	
	\Statex \textbf{Stage 1: Interaction-free Multi-granular Recall}
	\State Encode $x_q$ into global text embedding $t_q$ and textual tokens $T_q$
	\For{each image $I \in \mathcal{G}$}
	\State Compute or load the interaction-free multi-granular recall embedding $\mathbf{v}_1(I)$
	\EndFor
	\State $\mathcal{C} \leftarrow \mathrm{TopK}(\{\cos(t_q, \mathbf{v}_1(I))\}_{I \in \mathcal{G}}, K)$
	
	\Statex \textbf{Stage 2: Text-guided Cross-modal Reranking with BTIB}
	\For{each candidate image $I \in \mathcal{C}$}
	\State Construct coarse and fine visual tokens $\mathbf{H}^c(I)$ and $\mathbf{H}^f(I)$
	\State Compute $\mathbf{v}_2, \mathbf{v}_3 = \mathrm{BTIB}(\mathbf{H}^c(I), \mathbf{H}^f(I), T_q)$
	\State $s(I) \leftarrow (1-\lambda)\cos(t_q, \mathbf{v}_2) + \lambda \cos(t_q, \mathbf{v}_3)$
	\EndFor
	
	\State \Return candidates in $\mathcal{C}$ ranked by $s(I)$
\end{algorithmic}
\end{algorithm}
Algorithm~\ref{alg:ftf} summarizes the overall inference procedure of the proposed FTF framework.
The algorithm is written for T2I retrieval for clarity. For I2T retrieval, the same candidate filtering and reranking principle is applied by exchanging the query and target gallery, where each candidate text is paired with the query image and BTIB uses the textual tokens of the candidate sentence to refine the visual tokens of the query image before reranking.

\subsection{Interaction-Free Multi-Granular Recall}

In the interaction-free multi-granular recall stage, each RS image $I$ is encoded into a multi-granular visual embedding $\mathbf{v}_1$ by integrating coarse scene-level semantics and fine patch-level details. These gallery embeddings can be pre-computed before inference, allowing candidate selection to be performed through lightweight embedding-level similarity with the textual query embedding $t_q$. In this way, the recall stage avoids full-gallery token-level text--image interaction while preserving multi-granular visual semantics for efficient top-$K$ candidate selection.

Formally, a shared CLIP visual encoder first extracts a base patch-token grid from image $I$. Coarse and fine regional tokens are then constructed using window sets $\mathcal{P}_c=\{3\!\times\!3,5\!\times\!5\}$ and $\mathcal{P}_f=\{7\!\times\!7,9\!\times\!9\}$, respectively. These values denote average-pooling window sizes on the visual token grid, rather than input-pixel patch sizes or numbers of grid partitions. The two branches share the same visual backbone and differ only in their token-grid window scales. For window $w$ in branch $g$, the token sequence is
\begin{equation}
	\mathbf{H}^{g,w} = \mathrm{WindowPool}\!\left(E_{\mathrm{vis}}^{\mathrm{CLIP}}(I),w\right),
	\quad w\in\mathcal{P}_g,\; g\in\{c,f\}.
	\label{eq:multigranular_tokens}
\end{equation}
Tokens from all windows in the same branch are concatenated as $\mathbf{H}^{g}=\mathop{\Vert}_{w\in\mathcal{P}_g}\mathbf{H}^{g,w}$ for subsequent aggregation and reranking.

For each scale, mean and max pooling are concatenated to retain complementary distributional and salient-region information:
\begin{equation}
	\mathbf{p}^{g,w}=\left[
	\mathrm{MeanPool}(\mathbf{H}^{g,w})\,\Vert\,
	\mathrm{MaxPool}(\mathbf{H}^{g,w})\right].
	\label{eq:scale_pooling}
\end{equation}
Layer normalization is applied to the tokens before pooling and to each pooled scale feature. The four scale features are concatenated and mapped to the shared $d$-dimensional embedding space through two bias-free linear layers with a GELU activation:
\begin{equation}
	\mathbf{v}_1 = \mathrm{norm}_2\!\left(
	\mathbf{W}_r^{(2)}\,\mathrm{GELU}\!\left(
	\mathbf{W}_r^{(1)}
	\mathop{\Vert}_{g\in\{c,f\},\,w\in\mathcal{P}_g}\mathbf{p}^{g,w}
	\right)\right),
	\label{eq:recall_embedding}
\end{equation}
where $\Vert$ denotes feature concatenation and $\mathrm{norm}_2(\cdot)$ denotes $\ell_2$ normalization. The final projection $\mathbf{W}_r^{(2)}$ maps the fused visual feature to dimension $d$, ensuring that $\mathbf{v}_1$ and the global text embedding have the same dimensionality for cosine similarity.
Given a textual query embedding $t_q$, the interaction-free multi-granular recall stage efficiently retrieves the top-$K$ candidate set:
\begin{equation}
	\mathcal{C} = \mathrm{TopK}\!\left(\left\{ \cos\!\left(t_q, \mathbf{v}_1(I)\right) \right\}_{I \in \mathcal{G}}, K \right),
	\label{eq:topk_candidates}
\end{equation}
where $\mathcal{C}$ is the top-$K$ candidate set passed to the text-guided cross-modal reranking stage.  

This design ensures that $\mathbf{v}_1$ captures both global structures and local details, enabling efficient candidate selection using approximate nearest neighbor search while avoiding full-gallery token-level text--image interaction.

\subsection{Text-Guided Cross-Modal Reranking with BTIB}
 
After interaction-free multi-granular recall, the top-$K$ candidate set is selected using the similarity between the query text embedding and the pre-computed recall embedding $\mathbf{v}_1$. 
Recent studies have shown that structured text--visual interaction can improve fine-grained multimodal matching by guiding textual cues to focus on different semantic patterns in visual features~\cite{24eccvOTSeg}. 
Inspired by this observation, text-guided cross-modal reranking with BTIB refines the candidate ranking by introducing textual information through parameter-free token interaction.

For each candidate image, its coarse and fine tokens, $\mathbf{H}^c$ and $\mathbf{H}^f$, interact with the textual token embeddings $T$ to produce text-guided coarse and fine embeddings, $\mathbf{v}_2$ and $\mathbf{v}_3$:
\begin{equation}
	\mathbf{v}_2, \mathbf{v}_3 = \mathrm{BTIB}(\mathbf{H}^c, \mathbf{H}^f, T).
	\label{eq:btib_outputs}
\end{equation}

The core mechanism of BTIB is a bidirectional matching process. Let $\bar{T}$ and $\bar{\mathbf{H}}$ denote token-wise $\ell_2$-normalized text and visual features. The similarity matrix is
\begin{equation}
	S = \frac{\bar{T}\bar{\mathbf{H}}^\top}{\tau} \in \mathbb{R}^{N_t \times N_v},
	\label{eq:similarity_matrix}
\end{equation}
where $T \in \mathbb{R}^{N_t \times d}$ denotes valid, non-padding textual token embeddings, $\mathbf{H} \in \{\mathbf{H}^c, \mathbf{H}^f\}$ denotes coarse or fine image tokens corresponding to $\mathbf{v}_2$ or $\mathbf{v}_3$, and $\tau$ is a temperature hyperparameter.  

\textit{Text-conditioned visual normalization:} for each textual token, we apply a softmax over all visual tokens to obtain its visual-token assignment:
\begin{equation}
	A_{ij} = \frac{\exp(S_{ij})}{\sum_{k=1}^{N_v} \exp(S_{ik})}, \quad i=1,\dots,N_t,\; j=1,\dots,N_v,
	\label{eq:text_side_norm}
\end{equation}

\textit{Visual-token balancing normalization:} we also apply a softmax over all textual tokens for each visual token:
\begin{equation}
	B_{ij} = \frac{\exp(S_{ij})}{\sum_{l=1}^{N_t} \exp(S_{lj})}, \quad i=1,\dots,N_t,\; j=1,\dots,N_v.
	\label{eq:image_side_norm}
\end{equation}
The two directional assignments are combined through element-wise multiplication and normalized over the visual dimension:
\begin{equation}
	M=A\odot B, \qquad
	W_{ij}=\frac{M_{ij}}{\max\!\left(\sum_{k=1}^{N_v}M_{ik},\epsilon\right)},
	\label{eq:balanced_correspondence}
\end{equation}
where $\odot$ denotes element-wise multiplication and $\epsilon$ is a small numerical constant. This balanced correspondence emphasizes token pairs that are simultaneously distinctive from both the textual and visual perspectives.

The balanced assignments aggregate the candidate's visual values for each valid textual token:
\begin{equation}
	\mathbf{Z} = W\mathbf{H} \in \mathbb{R}^{N_t \times d},
	\label{eq:text_guided_visual_tokens}
\end{equation}
and the rows are averaged and $\ell_2$-normalized to generate coarse and fine reranking embeddings:
\begin{equation}
	\mathbf{v}_2 = \mathrm{norm}_2\!\left(\frac{1}{N_t}\sum_{i=1}^{N_t}\mathbf{Z}^{c}_{i}\right), \quad
	\mathbf{v}_3 = \mathrm{norm}_2\!\left(\frac{1}{N_t}\sum_{i=1}^{N_t}\mathbf{Z}^{f}_{i}\right),
	\label{eq:pool_v2_v3}
\end{equation}
The text-guided coarse and fine embeddings $\mathbf{v}_2$ and $\mathbf{v}_3$ are jointly optimized with the recall embedding $\mathbf{v}_1$ using the I$^2$M Loss. During reranking, their similarities to the query are combined to score each candidate:
\begin{equation}
	s(I) = (1-\lambda)\cos(t_q, \mathbf{v}_2) + \lambda \cos(t_q, \mathbf{v}_3),
	\label{eq:score_image}
\end{equation}
The candidate images in $\mathcal{C}$ are ranked according to $s(I)$. BTIB is a parameter-free interaction module that establishes correspondences between textual and visual tokens through bidirectional normalization and aggregates query-relevant coarse- and fine-grained visual information to refine candidate-level semantic matching.

\subsection{Inter- and Intra-Modal Loss}

Recent studies have shown that contrastive learning is an effective method for multimodal representation learning and cross-modal retrieval, and the InfoNCE~\cite{21icmlCLIPBASE} objective has been widely used to align matched visual--text pairs while separating non-matching pairs~\cite{22neurocomputingCLIP4Clip,25aclMMRet}. 
Motivated by this line of research, to jointly optimize the recall embedding $\mathbf{v}_1$ and the reranking embeddings $\mathbf{v}_2$ and $\mathbf{v}_3$ under a unified objective, we design an improved Inter- and Intra-Modal (I$^2$M) Loss. Specifically, the proposed loss combines an inter-modal alignment term to enforce cross-modal consistency with an intra-modal structural consistency term to preserve the relative structure of embeddings within each modality:
\begin{equation}
	\mathcal{L}_{\mathrm{I^2M}} = \mathcal{L}_{\mathrm{inter}} + \beta \mathcal{L}_{\mathrm{intra}},
	\label{eq:i2m_loss}
\end{equation}
where $\beta$ balances the two terms.

\textit{Inter-modal alignment.}
For a minibatch of $B$ image--text pairs, the inter-modal term aligns the global text embedding $\mathbf{t}_i$ with the recall and reranking representations of every image in the minibatch. The recall representation $\mathbf{v}_{1,j}$ depends only on image $I_j$, whereas the two BTIB representations are conditioned on both image $I_j$ and the textual tokens $T_i$ of the current query:
\begin{equation}
	\mathbf{v}_{1,j}=f_{\mathrm{recall}}(I_j), \qquad
	(\mathbf{v}_{2,ij},\mathbf{v}_{3,ij})=
	f_{\mathrm{BTIB}}(\mathbf{H}^{c}_j,\mathbf{H}^{f}_j,T_i).
	\label{eq:pair_conditioned_embeddings}
\end{equation}
Accordingly, the pairwise similarities are computed as
\begin{equation}
	\begin{aligned}
		s_{ij}^{(1)} &= \bar{\mathbf{t}}_i^\top \bar{\mathbf{v}}_{1,j},\\
		s_{ij}^{(k)} &= \bar{\mathbf{t}}_i^\top \bar{\mathbf{v}}_{k,ij},
		\quad k\in\{2,3\},
	\end{aligned}
	\label{eq:sim_kij}
\end{equation}
where the overbar denotes $\ell_2$ normalization. Thus, the reranking branches use the same query-conditioned image--text pairing during training and inference.  
We then adopt a sigmoid-based alignment objective:
\begin{equation}
	\mathcal{L}_{\mathrm{inter}}^{(k)} =
	\frac{1}{B^2}\sum_{i=1}^{B}\sum_{j=1}^{B}
	\log\left(1+\exp\left(-y_{ij}\, s_{ij}^{(k)}/\tau\right)\right),
	\label{eq:inter_loss_k}
\end{equation}
where $y_{ij}=1$ if $i=j$, and $y_{ij}=-1$ otherwise, and $\tau$ is a temperature hyperparameter.  
The overall inter-modal loss is defined as
\begin{equation}
	\begin{aligned}
		\mathcal{L}_{\mathrm{inter}}
		&= \alpha_1 \mathcal{L}_{\mathrm{inter}}^{(1)}
		+ \alpha_2 \mathcal{L}_{\mathrm{inter}}^{(2)}
		+ \alpha_3 \mathcal{L}_{\mathrm{inter}}^{(3)},
	\end{aligned}
	\label{eq:inter_loss}
\end{equation}
where $\alpha_1$, $\alpha_2$, and $\alpha_3$ control the contributions of the recall and reranking embeddings.  
This term encourages $\mathbf{v}_1$, $\mathbf{v}_2$, and $\mathbf{v}_3$ to maintain consistent semantic alignment with the paired text.

\textit{Intra-modal structural consistency.}
Although $\mathbf{v}_2$ and $\mathbf{v}_3$ are refined by text-guided cross-modal reranking with BTIB, they should preserve the intrinsic neighborhood structure established by the interaction-free multi-granular recall embedding $\mathbf{v}_1$. For the intra-modal term, $\mathbf{v}_{k,i}$ denotes the positive pair-conditioned representation $\mathbf{v}_{k,ii}$ for $k\in\{2,3\}$. We first $\ell_2$-normalize all representations as $\bar{\mathbf{v}}_{k,i}=\mathbf{v}_{k,i}/\lVert\mathbf{v}_{k,i}\rVert_2$ and construct a batch-level similarity graph based on $\mathbf{v}_1$:
\begin{equation}
	a_{ij} = \bar{\mathbf{v}}_{1,i}^\top \bar{\mathbf{v}}_{1,j}.
	\label{eq:aij_v1}
\end{equation}

For each sample $i$, we select its top-$M$ neighbors $\mathcal{N}_M(i)$ according to $a_{ij}$, excluding the sample itself, and define the normalized neighborhood weights:
\begin{equation}
	g_{ij} =
	\frac{\exp(a_{ij}/\sigma)}
	{\sum_{m \in \mathcal{N}_M(i)} \exp(a_{im}/\sigma)},
	\quad j \in \mathcal{N}_M(i),
	\label{eq:gij}
\end{equation}
where $\sigma$ is a temperature hyperparameter. The neighborhood weights are not detached: gradients propagate through the selected similarities and target embeddings, while the discrete top-$M$ index selection itself is treated as fixed within each forward pass.  
The intra-modal loss then enforces $\mathbf{v}_2$ and $\mathbf{v}_3$ to preserve this local structure:
\begin{equation}
	\begin{aligned}
		\mathcal{L}_{\mathrm{intra}}^{(k)}
		&= \frac{1}{B}\sum_{i=1}^{B}\sum_{j \in \mathcal{N}_M(i)}
		g_{ij}\left(1-\bar{\mathbf{v}}_{k,i}^\top \bar{\mathbf{v}}_{k,j}\right),
		\quad k \in \{2,3\}.
	\end{aligned}
	\label{eq:intra_loss_k}
\end{equation}

The final intra-modal term is
\begin{equation}
	\mathcal{L}_{\mathrm{intra}} =
	\mathcal{L}_{\mathrm{intra}}^{(2)} + \mathcal{L}_{\mathrm{intra}}^{(3)}.
	\label{eq:intra_loss}
\end{equation}

By jointly optimizing inter-modal alignment and intra-modal structural consistency, I$^2$M Loss encourages the recall and reranking embeddings to be semantically aligned with the query while maintaining consistent multi-granular representation structure.

\section{Experiments}
\subsection{Experimental Settings}

\textbf{Datasets.} We conduct experiments on the public RS image--text retrieval benchmark RSITMD~\cite{22tgrsAMFMN}. This dataset consists of paired RS images and textual descriptions, and exhibits high spatial resolution and complex background characteristics. To further evaluate generalization, we additionally include RSICD~\cite{18tgrsRSICD} and UCM~\cite{16citsUCMCaptions}, two widely used RS image--text retrieval benchmarks with different data scales, scene distributions, and caption granularities.

\textbf{Settings.} We initialize both the visual and textual branches from pretrained CLIP ViT-B/32~\cite{21icmlCLIPBASE}. The shared CLIP visual encoder extracts base patch tokens from which the coarse and fine window branches are constructed, while the CLIP text encoder produces token-level and sentence-level representations of each description. The encoders, projection layers, and other trainable components are adapted on the training split of each target RS image--text retrieval dataset using the proposed objective. All experiments are conducted on a single NVIDIA A100 GPU with 40GB memory. Images are resized to $224\times224$, the maximum text length is set to 77 tokens, and both image and text embeddings are projected to 512 dimensions. We train FTF for 30 epochs with a batch size of 128 using AdamW, with an initial learning rate of $1\times10^{-5}$ and a weight decay of 0.05. At inference time, the system follows the FTF two-stage retrieval pipeline: it first performs interaction-free multi-granular recall over the full gallery and then applies text-guided cross-modal reranking with BTIB only to the top-$K$ candidate set. The same procedure is used for both I2T and T2I retrieval, with candidate filtering performed over the corresponding target gallery before reranking. Unless otherwise specified, the candidate size $K$ is selected from $\{50,100,200\}$. All datasets follow the data partitioning protocol of GaLR~\cite{22tgrsGaLR}.

\textbf{Evaluation metrics.} We report Recall@K for both image-to-text retrieval (I2T) and text-to-image retrieval (T2I), where $K\in\{1,5,10\}$, and compute mean Recall (mR) as the average of these six values. Let $\mathcal{Q}$ denote the query set, $\mathrm{TopK}(q)$ the top-$K$ retrieved items for query $q$, and $\mathcal{G}(q)$ the set of its ground-truth matched items. Recall@$K$ is defined as
\begin{equation}
	\mathrm{R@}K
	=
	\frac{1}{|\mathcal{Q}|}
	\sum_{q \in \mathcal{Q}}
	\mathbf{1}\!\left(\mathcal{G}(q)\cap \mathrm{TopK}(q)\neq \emptyset\right),
	\label{eq:recall_at_k}
\end{equation}
where $\mathbf{1}(\cdot)$ equals 1 if the condition is true and 0 otherwise. To evaluate retrieval efficiency, we further report latency (Lat.), defined as the average inference time per query, and throughput (Thr.), defined as the number of queries processed per second.

\textbf{Compared methods.} We compare FTF with 20 representative methods, including LW-MCR~\cite{21tgrsLW-MCR}, AMFMN~\cite{22tgrsAMFMN}, KAMCL~\cite{23tgrsKAMCL}, GaLR~\cite{22tgrsGaLR}, PIR~\cite{23acmmmPIR}, FAAMI~\cite{23sensorsFAAMI}, MTGFE~\cite{23RemoteSensingMTGFE}, MSCMAT~\cite{24tgrsMSCMAT}, IERR~\cite{24tgrsIERR}, VGSGN~\cite{24tgrsVGSGN}, SIPN~\cite{25tgrsSIPN}, FSSN~\cite{25tgrsFSSN}, CLIP-Base~\cite{21icmlCLIPBASE}, PETL~\cite{23tgrsPETL}, SkyScript~\cite{24aaaiskysrcipt}, GRAFT~\cite{24iclrGRAFT}, RemoteCLIP~\cite{24tgrsremoteclip}, GeoRSCLIP~\cite{24tgrsGeoRSCLIP}, EBAKER~\cite{24acmmmEBAKER}, and EAPA~\cite{25acmmmEAPA}. For methods based on pretrained vision--language models, we report fine-tuned retrieval results.

\subsection{Comparison with State-of-the-Art Methods}

\begin{table}[t]
	\caption{Comparison of state-of-the-art methods on RSITMD. For FTF, Full denotes full-gallery BTIB reranking without candidate pruning, while $K>0$ denotes interaction-free multi-granular recall followed by text-guided cross-modal reranking with BTIB over the top-$K$ candidate set. }
	\label{tab:rsitmd_sota_eff}
	\centering
	\footnotesize
	\renewcommand{\arraystretch}{1.1}
	\resizebox{\textwidth}{!}{%
		\begin{tabular}{lcccccccccc}
			\toprule
			Method & Year & \multicolumn{3}{c}{I2T} & \multicolumn{3}{c}{T2I} & mR$\uparrow$ & Lat.\,(ms)$\downarrow$ & Thr.\,(q/s)$\uparrow$ \\
			\cmidrule(lr){3-5}
			\cmidrule(lr){6-8}
			& & R@1 & R@5 & R@10 & R@1 & R@5 & R@10 & & & \\
			\midrule
			
			LW-MCR & 2022 & 9.43 & 26.18 & 37.87 & 8.52 & 34.13 & 53.80 & 28.32 & 172.8 & 5.8 \\
			AMFMN  & 2022 & 10.31 & 29.21 & 37.93 & 9.85 & 33.30 & 52.26 & 28.81 & 211.3 & 4.7 \\
			GaLR   & 2022 & 14.69 & 32.20 & 41.84 & 10.68 & 36.90 & 52.44 & 31.46 & 246.8 & 4.1 \\
			KAMCL  & 2023 & 14.95 & 29.47 & 44.39 & 9.76 & 36.41 & 55.48 & 31.74 & 185.6 & 5.4 \\
			PIR    & 2023 & 17.54 & 40.50 & 52.55 & 12.71 & 41.14 & 63.55 & 38.00 & 198.5 & 5.0 \\
			FAAMI  & 2023 & 16.39 & 35.40 & 48.97 & 12.22 & 41.64 & 59.46 & 35.68 & 202.7 & 4.9 \\
			MTGFE  & 2023 & 18.23 & 40.81 & 53.00 & 16.74 & 48.42 & 67.09 & 40.72 & 224.6 & 4.5 \\
			MSCMAT & 2024 & 22.85 & 43.26 & 55.31 & 15.31 & 47.39 & 65.37 & 41.58 & 236.9 & 4.2 \\
			IERR   & 2024 & 28.05 & 48.53 & 61.88 & 23.78 & 56.45 & 73.10 & 48.63 & 258.4 & 3.9 \\
			VGSGN  & 2024 & 13.57 & 34.94 & 49.88 & 13.52 & 43.02 & 63.55 & 36.41 & 190.6 & 5.2 \\
			SIPN   & 2025 & 23.65 & 45.70 & 58.29 & 17.20 & 50.01 & 66.88 & 43.62 & 215.8 & 4.6 \\
			FSSN   & 2025 & \underline{31.11} & \underline{52.97} & 65.45 & 25.99 & 57.57 & \textbf{75.74} & 51.47 & 238.0 & 4.2 \\
			\midrule
			
			CLIP-Base  & 2021 & 9.36 & 21.26 & 32.25 & 8.09 & 26.99 & 40.27 & 23.04 & 145.5 & 6.9 \\
			PETL       & 2023 & 23.35 & 47.05 & 60.72 & 19.76 & 49.78 & 69.00 & 44.94 & 157.6 & 6.3 \\
			SkyScript  & 2024 & 11.42 & 35.31 & 53.49 & 14.44 & 32.43 & 40.39 & 31.25 & 152.9 & 6.5 \\
			GRAFT      & 2024 & 25.33 & 45.33 & 58.43 & 25.54 & 53.01 & 61.33 & 44.83 & 163.8 & 6.1 \\
			RemoteCLIP & 2024 & 27.64 & 51.31 & 66.49 & 21.58 & 55.91 & 72.95 & 49.31 & 150.7 & 6.6 \\
			GeoRSCLIP  & 2024 & 29.64 & 51.52 & 63.42 & 23.14 & 56.68 & 74.46 & 49.81 & 153.5 & 6.5 \\
			EBAKER     & 2024 & \textbf{33.85} & \textbf{54.31} & \textbf{68.72} & \textbf{28.37} & \textbf{60.38} & \underline{75.51} & \textbf{53.52} & 166.4 & 6.0 \\
			EAPA       & 2025 & 30.17 & 52.56 & 63.51 & 24.64 & 58.12 & 72.32 & 50.22 & 160.5 & 6.2 \\
			\midrule
			
			\textbf{FTF (Full)}   
			& 2026 
			& 30.97 
			& 52.18 
			& \underline{66.78} 
			& 27.39
			& \underline{59.76} 
			& 73.30 
			& \underline{51.73} 
			& 295.4 
			& 3.4 \\
			\textbf{FTF (K=50)}  & 2026 & 27.44 & 49.77 & 63.18 & 24.91 & 54.66 & 71.94 & 48.65 & \textbf{12.4} & \textbf{80.6} \\
			\textbf{FTF (K=100)} & 2026 & 29.83 & 50.69 & 65.64 & 26.42 & 58.21 & 72.15 & 50.49 & \underline{22.9} & \underline{43.7} \\
			\textbf{FTF (K=200)} & 2026 & 30.03 & 52.78 & 65.77 & \underline{27.76} & 58.68 & 73.80 & 51.47 & 39.6 & 25.3 \\
			\bottomrule
		\end{tabular}%
	}
\end{table}

\begin{figure}[t]
	\centering
	\includegraphics[width=\linewidth,height=0.45\textheight,keepaspectratio]{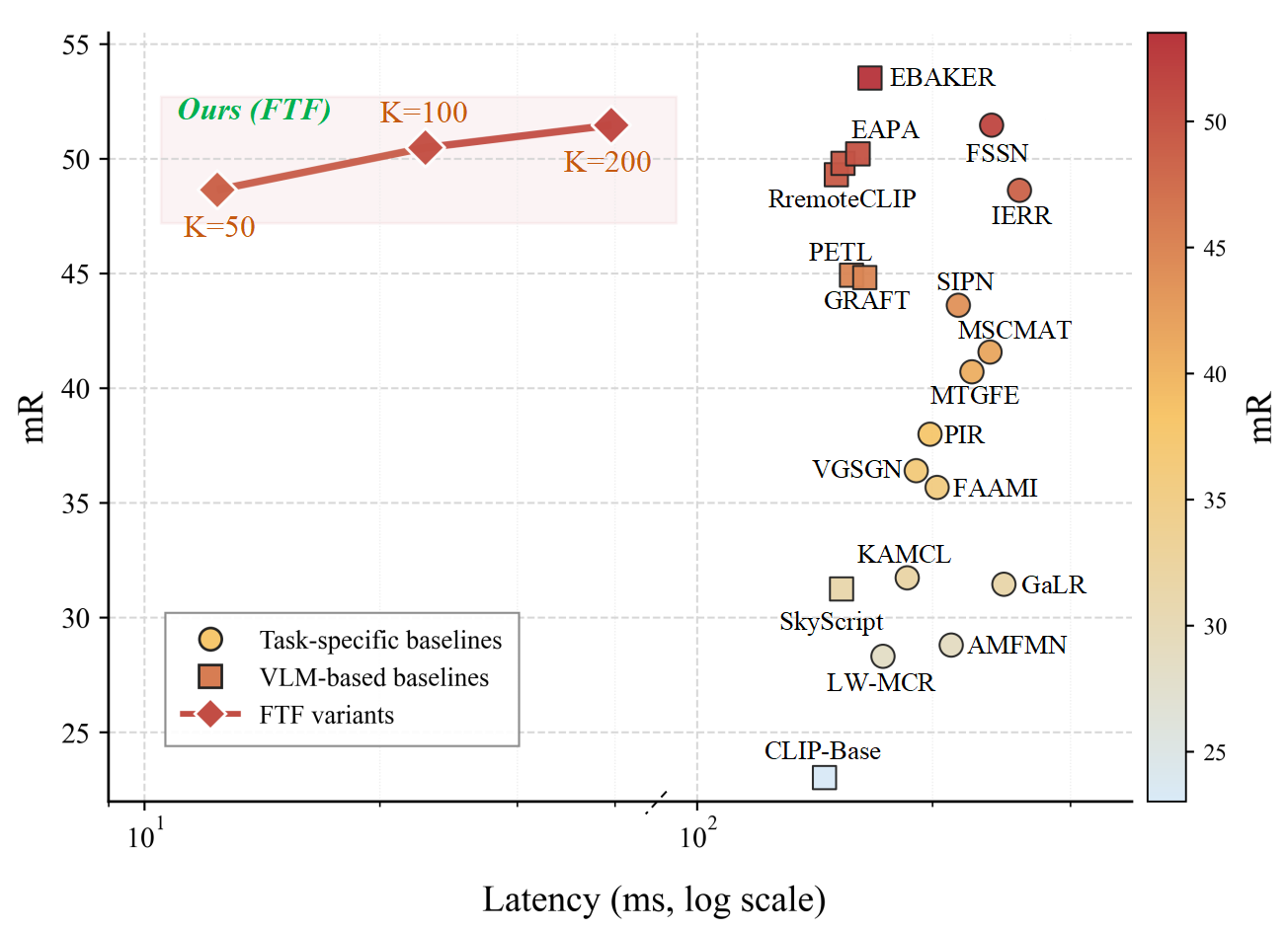}
	\caption{Accuracy--efficiency comparison on RSITMD using the methods reported in Table~\ref{tab:rsitmd_sota_eff}. FTF provides favorable operating points under different candidate sizes $K$.}
	\Description{Scatter plot comparing mean recall and inference latency on RSITMD; FTF variants provide low-latency operating points while retaining competitive mean recall.}
	\label{fig:intro_tradeoff}
\end{figure}

Table~\ref{tab:rsitmd_sota_eff} reports the main comparison results on RSITMD, and Fig.~\ref{fig:intro_tradeoff} visualizes the corresponding accuracy--efficiency distribution. Overall, FTF achieves a favorable accuracy--efficiency balance across different candidate sizes $K$. The full-gallery BTIB reranking setting, FTF (Full), obtains an mR of 51.73, which is competitive with strong task-specific and VLM-based baselines. Although EBAKER achieves the highest mR, FTF (Full) remains close to the strongest results and outperforms several representative methods such as IERR, SIPN, FSSN, RemoteCLIP, GeoRSCLIP, and EAPA. This indicates that text-guided cross-modal reranking with BTIB is effective when applied to the complete gallery. However, full-gallery BTIB reranking also introduces relatively high latency, which motivates the FTF two-stage retrieval framework.

When candidate pruning is enabled, FTF provides a flexible balance between retrieval accuracy and inference efficiency. With candidate size $K=50$, FTF achieves the lowest latency (12.4 ms) and the highest throughput (80.6 q/s) among all compared methods, while maintaining a competitive mR of 48.65. Increasing the candidate size gradually recovers retrieval accuracy: FTF ($K=100$) achieves higher mR than RemoteCLIP and GeoRSCLIP, and FTF ($K=200$) reaches an mR of 51.47, which is close to FTF (Full) and comparable to the strong task-specific baseline FSSN, while requiring only 39.6 ms latency. These results show that candidate size $K$ serves as an explicit control knob: smaller $K$ values favor real-time retrieval efficiency, while larger $K$ values recover fine-grained matching accuracy.

Compared with VLM-based methods, FTF is particularly useful under practical inference constraints. Rather than pursuing the highest single-point mR, FTF provides adjustable operating points with substantially lower latency. For example, FTF ($K=50$) is substantially faster than all methods with available efficiency results, and FTF ($K=200$) achieves higher mR than RemoteCLIP, GeoRSCLIP, and EAPA. This demonstrates that the proposed framework does not rely on heavy full-gallery token-level text--image interaction; instead, it restricts text-guided cross-modal reranking with BTIB to the top-$K$ candidate set, thereby maintaining competitive retrieval performance while substantially reducing inference cost.

\subsection{Generalization and Efficiency on Additional Datasets}

To evaluate the generalization ability of the proposed FTF, we further conduct experiments on two additional RS image--text retrieval benchmarks, RSICD and UCM. RSICD contains more images and more diverse scene categories, while UCM is a compact benchmark with different caption and category distributions. These complementary datasets allow us to assess whether the accuracy--efficiency balance of FTF is consistent across different data scales and retrieval settings.

\begin{table}[t]
	\caption{Comparison of retrieval performance and efficiency on RSICD and UCM.}
	\label{tab:additional_datasets_eff}
	\centering
	\footnotesize
	\renewcommand{\arraystretch}{1.08}
	\resizebox{\textwidth}{!}{%
		\begin{tabular}{llccccccccc}
			\toprule
			Dataset & Method & \multicolumn{3}{c}{I2T} & \multicolumn{3}{c}{T2I} & mR$\uparrow$ & Lat. (ms)$\downarrow$ & Thr. (q/s)$\uparrow$ \\
			\cmidrule(lr){3-5}
			\cmidrule(lr){6-8}
			& & R@1 & R@5 & R@10 & R@1 & R@5 & R@10 & & & \\
			\midrule
			
			\multirow{11}{*}{RSICD} & LW-MCR & 3.34 & 13.00 & 21.95 & 3.00 & 15.38 & 27.96 & 14.11 & 386.4 & 2.6 \\
			& AMFMN & 4.59 & 15.33 & 27.99 & 4.93 & 17.42 & 28.33 & 16.43 & 472.8 & 2.1 \\
			& PIR & 10.38 & 27.03 & 38.84 & 6.33 & 24.20 & 39.90 & 24.45 & 435.6 & 2.3 \\
			& RemoteCLIP & 16.79 & 34.94 & 48.25 & 13.91 & 40.79 & 56.58 & 35.21 & 333.5 & 3.0 \\
			& FSSN & \underline{18.94} & \underline{39.65} & \underline{51.71} & \underline{16.30} & 39.32 & 56.92 & 37.14 & 548.9 & 1.8 \\
			& EAPA & 18.24 & 38.74 & 50.62 & 15.66 & 41.24 & 58.04 & 37.09 & 347.1 & 2.9 \\
			& EBAKER & \textbf{20.12} & \textbf{41.32} & \textbf{53.76} & \textbf{17.42} & \textbf{43.58} & 56.96 & \textbf{38.86} & 352.4 & 2.8 \\
			& \textbf{FTF (Full)} & 18.17 & 37.86 & 50.44 & 15.38 & \underline{42.73} & \textbf{58.32} & \underline{37.15} & 680.4 & 1.5 \\
			& \textbf{FTF (K=50)} & 15.83 & 36.42 & 48.07 & 14.56 & 39.74 & 57.18 & 35.30 & \textbf{14.6} & \textbf{68.5} \\
			& \textbf{FTF (K=100)} & 17.23 & 36.72 & 49.46 & 14.48 & 41.37 & 57.52 & 36.13 & \underline{25.8} & \underline{38.8} \\
			& \textbf{FTF (K=200)} & 17.87 & 37.74 & 50.18 & 15.25 & 42.14 & \underline{58.16} & 36.89 & 42.3 & 23.6 \\
			
			\midrule
			
			\multirow{11}{*}{UCM} & LW-MCR & 14.06 & 39.87 & 59.14 & 9.59 & 44.07 & 74.42 & 40.19 & 76.8 & 13.0 \\
			& AMFMN & 14.80 & 51.04 & 67.02 & 14.64 & 51.74 & 80.35 & 46.60 & 93.5 & 10.7 \\
			& PIR & 12.43 & 47.98 & 75.25 & 13.76 & 53.08 & 91.13 & 48.94 & 88.6 & 11.3 \\
			& RemoteCLIP & 16.12 & 50.57 & \textbf{81.33} & 17.02 & 63.18 & \textbf{98.46} & 54.45 & 66.7 & 15.0 \\
			& FSSN & \textbf{23.18} & \textbf{64.27} & 81.07 & \textbf{21.06} & \textbf{66.19} & \underline{94.51} & \textbf{58.38} & 105.3 & 9.5 \\
			& EAPA & 21.16 & 61.54 & 80.42 & 20.13 & 64.39 & 94.12 & 56.96 & 67.4 & 14.8 \\
			& EBAKER & \underline{22.47} & \underline{63.18} & \underline{81.25} & 20.62 & \underline{65.73} & 94.09 & \underline{57.89} & 68.1 & 14.7 \\
			& \textbf{FTF (Full)} & 21.83 & 62.21 & 80.54 & \underline{20.78} & 65.37 & 93.91 & 57.44 & 130.8 & 7.6 \\
			& \textbf{FTF (K=50)} & 19.42 & 57.13 & 79.36 & 18.55 & 63.58 & 93.34 & 55.23 & \textbf{11.2} & \textbf{89.3} \\
			& \textbf{FTF (K=100)} & 19.86 & 60.73 & 79.58 & 20.18 & 63.71 & 94.22 & 56.38 & \underline{20.9} & \underline{47.8} \\
			& \textbf{FTF (K=200)} & 21.34 & 61.86 & 81.02 & 20.58 & 64.79 & 94.33 & 57.32 & 36.9 & 27.1 \\
			
			\bottomrule
		\end{tabular}%
	}
\end{table}

Table~\ref{tab:additional_datasets_eff} focuses on representative methods with comparable efficiency evaluation, covering early lightweight models, fine-grained task-specific models, and VLM-based models.
On RSICD, FTF (Full) achieves an mR of 37.15, slightly higher than FSSN (37.14) and RemoteCLIP (35.21). It obtains the best T2I R@10 and the second-best T2I R@5, indicating that full-gallery BTIB reranking remains effective on a larger and more diverse dataset. With candidate pruning, increasing the candidate size from $K=50$ to $K=200$ improves mR from 35.30 to 36.89 while keeping latency far lower than full-gallery BTIB reranking. On UCM, FTF (Full) reaches an mR of 57.44, close to the strongest baseline FSSN (58.38), and the two-stage variants preserve competitive retrieval accuracy under different candidate sizes $K$. Although the highest mR is obtained by EBAKER on RSICD and FSSN on UCM, FTF provides much lower latency under candidate pruning, showing a practical accuracy--efficiency balance across datasets.

To complement the absolute latency and throughput reported in Table~\ref{tab:additional_datasets_eff}, we visualize the relative speedup of each method by using RemoteCLIP as the dataset-specific reference.

\begin{figure}[!htbp]
	
	\centering
	\includegraphics[width=0.70\linewidth]{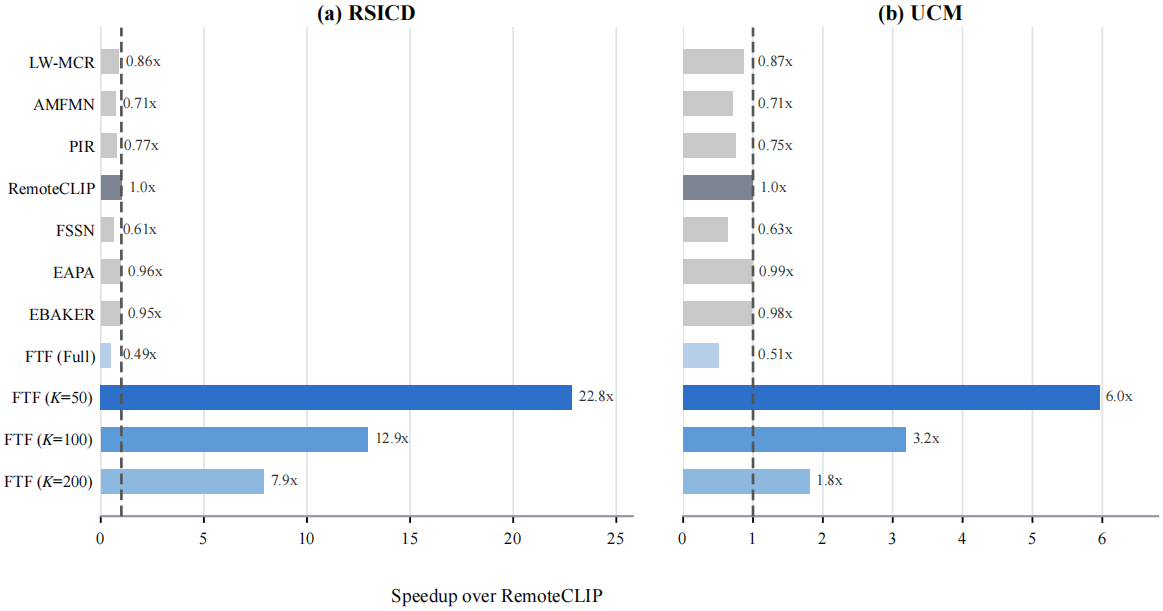}
	\caption{Relative inference speedup over RemoteCLIP on additional datasets. Speedup is computed as the latency of RemoteCLIP divided by the latency of each method on the same dataset. Gray bars denote compared baselines, and blue bars denote FTF variants under different candidate sizes $K$.}
	\Description{Bar chart of relative speedup over RemoteCLIP on RSICD and UCM, showing the largest gains for FTF with small candidate sets.}
	\label{fig:remoteclip_speedup}
\end{figure}

As shown in Fig.~\ref{fig:remoteclip_speedup}, FTF with candidate size $K=50$ provides clear inference acceleration over RemoteCLIP. Specifically, FTF ($K=50$) is 22.8$\times$ faster than RemoteCLIP on RSICD and 6.0$\times$ faster on UCM. The Full setting performs full-gallery BTIB reranking and is therefore slower, but it serves as a full-gallery reranking reference. In contrast, $K>0$ offers a practical accuracy--efficiency balance: smaller candidate sizes bring stronger speedups, while larger candidate sizes recover more retrieval accuracy.

\begin{figure}[!htbp]
	\centering
	\includegraphics[width=0.78\linewidth,height=0.25\textheight,keepaspectratio]{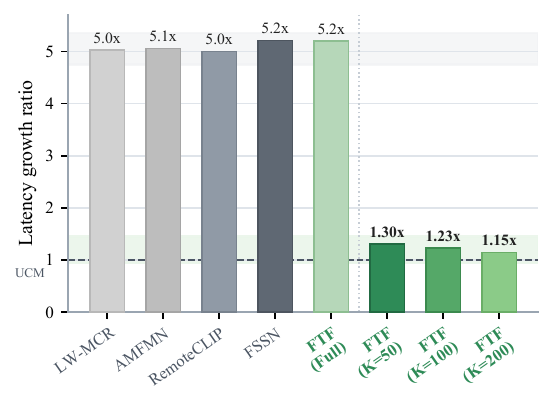}
	\caption{Latency growth ratio from UCM to RSICD. The growth ratio is computed as the latency on RSICD divided by the latency on UCM for each method. Smaller values indicate better scalability as the gallery size increases.}
	\Description{Bar chart of latency growth from UCM to RSICD; two-stage FTF variants grow substantially more slowly than full-gallery methods.}
	\label{fig:latency_scalability}
\end{figure}

Beyond relative speedup on each dataset, we further analyze how inference latency scales when the gallery size increases from UCM to RSICD. As shown in Fig.~\ref{fig:latency_scalability}, full-gallery matching methods exhibit clear latency growth, with the compared baselines increasing by about $5.0\times$--$5.2\times$. FTF (Full) follows a similar trend because BTIB reranking is applied to the complete gallery. In contrast, the two-stage FTF variants grow much more slowly, with only $1.30\times$, $1.23\times$, and $1.15\times$ latency growth for $K=50$, $K=100$, and $K=200$, respectively. This scalability advantage primarily comes from restricting text-guided cross-modal reranking with BTIB to a fixed-size top-$K$ candidate set, while interaction-free multi-granular recall efficiently scans the expanding gallery.

\subsection{Stage-1 Candidate Coverage}

To quantify the upper-bound constraint introduced by candidate filtering, we report the candidate coverage of the interaction-free multi-granular recall stage before text-guided cross-modal reranking with BTIB. Coverage@$K$ measures the percentage of queries whose ground-truth set has at least one matched item retained in the top-$K$ candidate set.

\begin{table}[t]
	\caption{Stage-1 candidate coverage before text-guided cross-modal reranking.}
	\label{tab:stage1_coverage}
	\centering
	\footnotesize
	\renewcommand{\arraystretch}{1.08}
	\begin{tabular}{llccc}
		\toprule
		Dataset & Direction & Cov@50 (\%) & Cov@100 (\%) & Cov@200 (\%) \\
		\midrule
		RSITMD & I2T & 90.6 & 94.7 & 97.6 \\
		RSITMD & T2I & 89.8 & 94.1 & 97.0 \\
		RSITMD & Avg. & 90.2 & 94.4 & 97.3 \\
		\midrule
		RSICD & I2T & 88.4 & 92.6 & 96.1 \\
		RSICD & T2I & 87.6 & 92.0 & 95.5 \\
		RSICD & Avg. & 88.0 & 92.3 & 95.8 \\
		\midrule
		UCM & I2T & 93.2 & 96.0 & 98.4 \\
		UCM & T2I & 92.5 & 95.6 & 98.0 \\
		UCM & Avg. & 92.9 & 95.8 & 98.2 \\
		\bottomrule
	\end{tabular}
\end{table}

As shown in Table~\ref{tab:stage1_coverage}, the interaction-free multi-granular recall stage preserves most ground-truth matches across all three datasets. The average Cov@200 reaches 97.3\%, 95.8\%, and 98.2\% on RSITMD, RSICD, and UCM, respectively, indicating that moderate candidate budgets provide sufficient candidate coverage for subsequent text-guided cross-modal reranking with BTIB.

\subsection{Ablation Study}

To verify the effectiveness of the proposed modules, we conduct ablation experiments on RSITMD under the setting of $K=100$. Specifically, we analyze three key components of FTF, including multi-granular learning, BTIB, and the I$^2$M Loss. For simplicity, we report I2T R@1, T2I R@1, and mR as representative metrics.

\begin{table}[t]
	\caption{Ablation study of FTF on RSITMD with $K=100$.}
	\label{tab:ablation}
	\centering
	\footnotesize
	\renewcommand{\arraystretch}{1.05}
	\begin{tabular}{lccc}
		\toprule
		Variant & I2T R@1 & T2I R@1 & mR \\
		\midrule
		Coarse only & 18.42 & 16.87 & 39.26 \\
		Fine only & 24.83 & 22.94 & 45.31 \\
		w/o BTIB & 27.96 & 25.21 & 46.67 \\
		w/o intra-modal term & 29.08 & 26.11 & 48.72 \\
		Full FTF & 29.83 & 26.42 & 50.49 \\
		\bottomrule
	\end{tabular}
\end{table}

Table~\ref{tab:ablation} reports the ablation results. First, the performance of \emph{Coarse only} and \emph{Fine only} is significantly lower than that of the full model, which confirms the necessity of multi-granular learning. In particular, \emph{Fine only} clearly outperforms \emph{Coarse only}, indicating that fine-grained local details are more important than coarse global structures for RS image--text retrieval. However, the best performance is achieved only when both branches are used together, showing that coarse and fine representations are complementary.

Second, removing BTIB leads to a clear performance drop, with mR decreasing from 50.49 to 46.67. This result demonstrates that the proposed parameter-free text-guided interaction is effective for refining semantic correspondence between RS imagery and textual descriptions. Third, removing the intra-modal term from the I$^2$M Loss also degrades performance, reducing mR from 50.49 to 48.72. This observation shows that, besides inter-modal alignment, preserving the neighborhood structure across multi-granular representations provides additional optimization benefits. 

\begin{table}[t]
		\centering
		\footnotesize
		\renewcommand{\arraystretch}{1.08}
		\caption{Comparison between BTIB and lightweight reranking alternatives on RSITMD with $K=100$. The reported parameters denote the additional parameters introduced by the reranking module.}
		\label{tab:btib_rerank_alternatives}
		\resizebox{\columnwidth}{!}{%
			\begin{tabular}{lcccccc}
				\toprule
				Reranking strategy & Params & I2T R@1 & T2I R@1 & mR$\uparrow$ & Lat. (ms)$\downarrow$ & Thr. (q/s)$\uparrow$ \\
				\midrule
				No reranking & 0 & 27.96 & 25.21 & 46.67 & \textbf{12.2} & \textbf{82.0} \\
				Score-level fusion & 0 & 26.87 & 24.61 & 48.12 & 15.8 & 63.3 \\
				Text-to-image softmax & 0 & 28.16 & 25.84 & 49.31 & 18.9 & 52.9 \\
				Lightweight cross-attention & 0.52M & \underline{29.31} & \underline{26.18} & \underline{50.21} & 33.4 & 29.9 \\
				\textbf{BTIB} & \textbf{0} & \textbf{29.83} & \textbf{26.42} & \textbf{50.49} & \underline{22.9} & \underline{43.7} \\
				\bottomrule
			\end{tabular}%
		}
	\end{table}

Table~\ref{tab:btib_rerank_alternatives} further compares BTIB with lightweight reranking alternatives. Compared with no reranking and score-level fusion, token-level text-guided interaction clearly improves retrieval accuracy. BTIB also outperforms the one-directional text-to-image softmax, indicating that the visual-token balancing normalization is necessary for alleviating token crowding and improving candidate-level alignment. Although lightweight cross-attention achieves competitive accuracy, it introduces extra parameters and higher latency, whereas BTIB obtains the best mR with no additional parameters and lower text-guided cross-modal reranking cost.

\subsection{Efficiency--Accuracy Balance Analysis}
 
\begin{figure}[!htb]
	\centering
	\includegraphics[width=\linewidth,height=0.27\textheight,keepaspectratio]{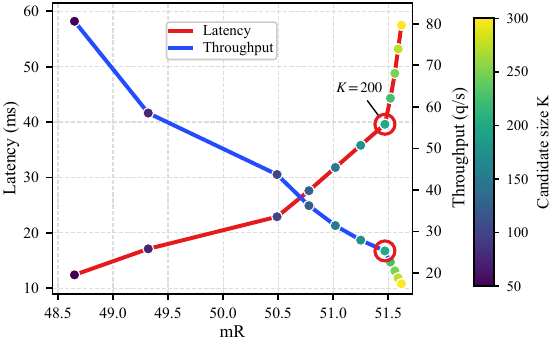}
\caption{Accuracy--efficiency balance of FTF under different candidate sizes $K$. The plot shows latency and throughput versus mean recall under different candidate sizes $K$. Point color encodes $K$, and the figure highlights the balance between retrieval accuracy and efficiency.}
	\Description{Dual-axis plot showing that mean recall rises with candidate size while latency increases and throughput decreases.}
	\label{fig:mR_tradeoff}
\end{figure}

To investigate the impact of candidate size $K$ on retrieval performance, we evaluated FTF with $K$ ranging from 50 to 300 on RSITMD.
Fig.~\ref{fig:mR_tradeoff} illustrates the balance between retrieval accuracy (mR) and efficiency metrics, including latency and throughput.
Each colored point corresponds to a specific candidate size $K$.

As shown in Fig.~\ref{fig:mR_tradeoff}, mR improves steadily as the candidate size increases from $K=50$ to $K=200$, and then gradually saturates, reaching 51.62 at $K=300$.
In contrast, latency increases almost monotonically from 12.4 ms to 57.5 ms, while throughput decreases from 80.6 q/s to 17.4 q/s.
This trend shows that larger candidate sets recover more fine-grained matching accuracy, but the marginal accuracy gain becomes small after $K=200$.
Therefore, candidate size $K$ provides an explicit accuracy--efficiency control knob, and $K=200$ offers a practical balance between competitive retrieval accuracy and moderate inference cost.

\subsection{Qualitative Analysis}

\begin{figure}[!htbp]
	\centering
	\includegraphics[width=0.86\linewidth,height=0.43\textheight,keepaspectratio]{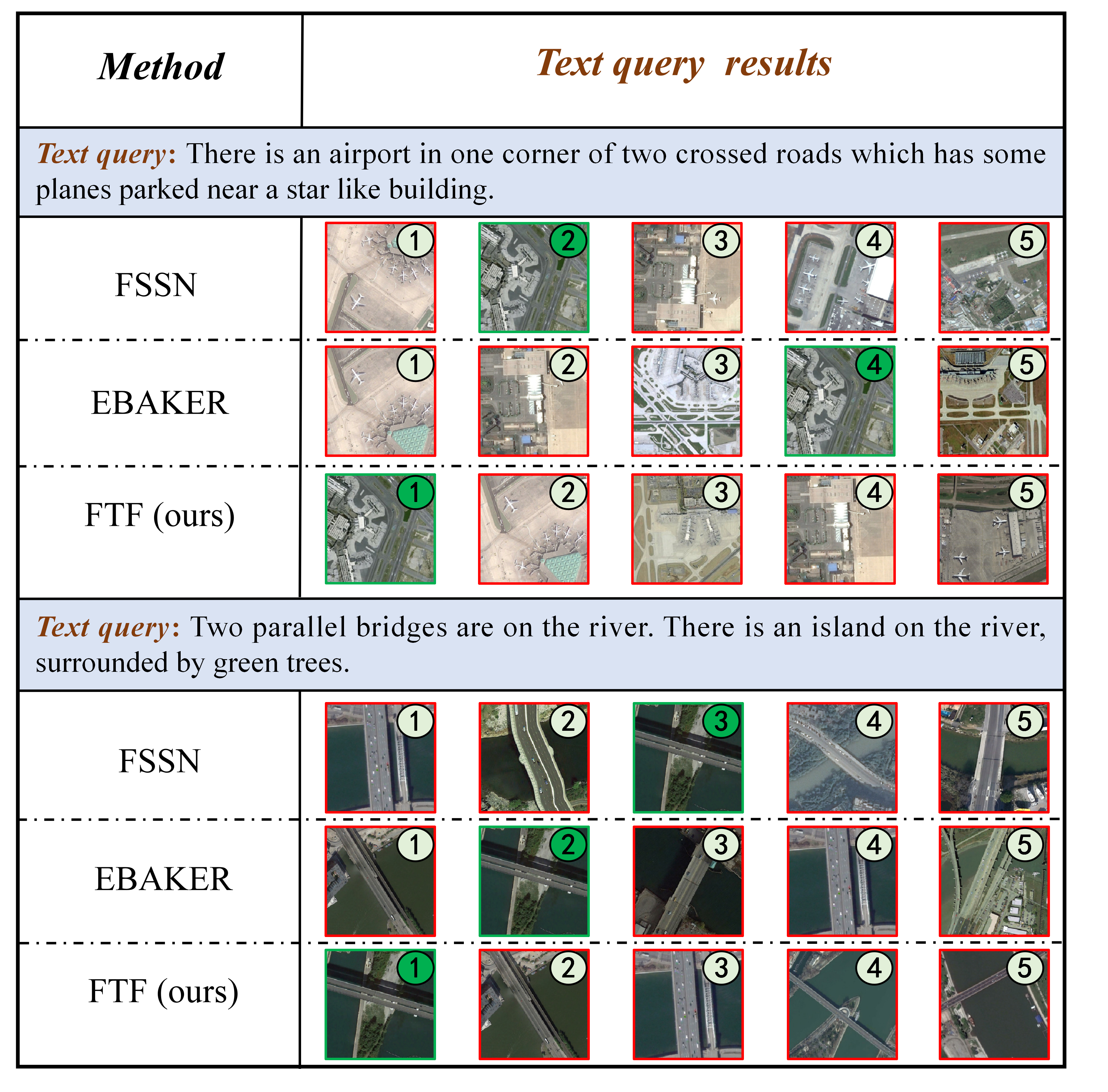}
	\caption{Qualitative comparisons with representative baselines. For each text query, we show the top-5 retrieved results of FSSN, EBAKER, and the proposed method. The numbers denote retrieval ranks, and ground-truth images are highlighted with green boxes. The proposed method retrieves results that are more consistent with both global semantics and fine-grained spatial details.}
	\Description{Top-five retrieval examples for airport and bridge queries compare FSSN, EBAKER, and FTF, with ground-truth images marked by green boxes.}
	\label{fig:baseline_compare}
\end{figure}

\textbf{Comparison with representative baselines.}
Fig.~\ref{fig:baseline_compare} shows qualitative comparisons with representative baselines under two challenging queries. For the airport query, FSSN and EBAKER retrieve semantically related airport scenes, but fail to rank the ground-truth image at the top because they do not fully capture the detailed spatial constraints. In contrast, the proposed method ranks the correct image at position 1, showing better alignment with both global semantics and fine-grained layout details. For the bridge query, a similar trend can be observed: although the baselines return visually related bridge scenes, their results only partially satisfy the query conditions, whereas our method again ranks the ground-truth image first. These examples show that the combination of coarse semantic understanding and fine-grained structural discrimination leads to more consistent retrieval results across queries requiring different levels of semantic detail.

\begin{figure}[!htbp]
	\centering
	\includegraphics[width=0.90\linewidth,height=0.36\textheight,keepaspectratio]{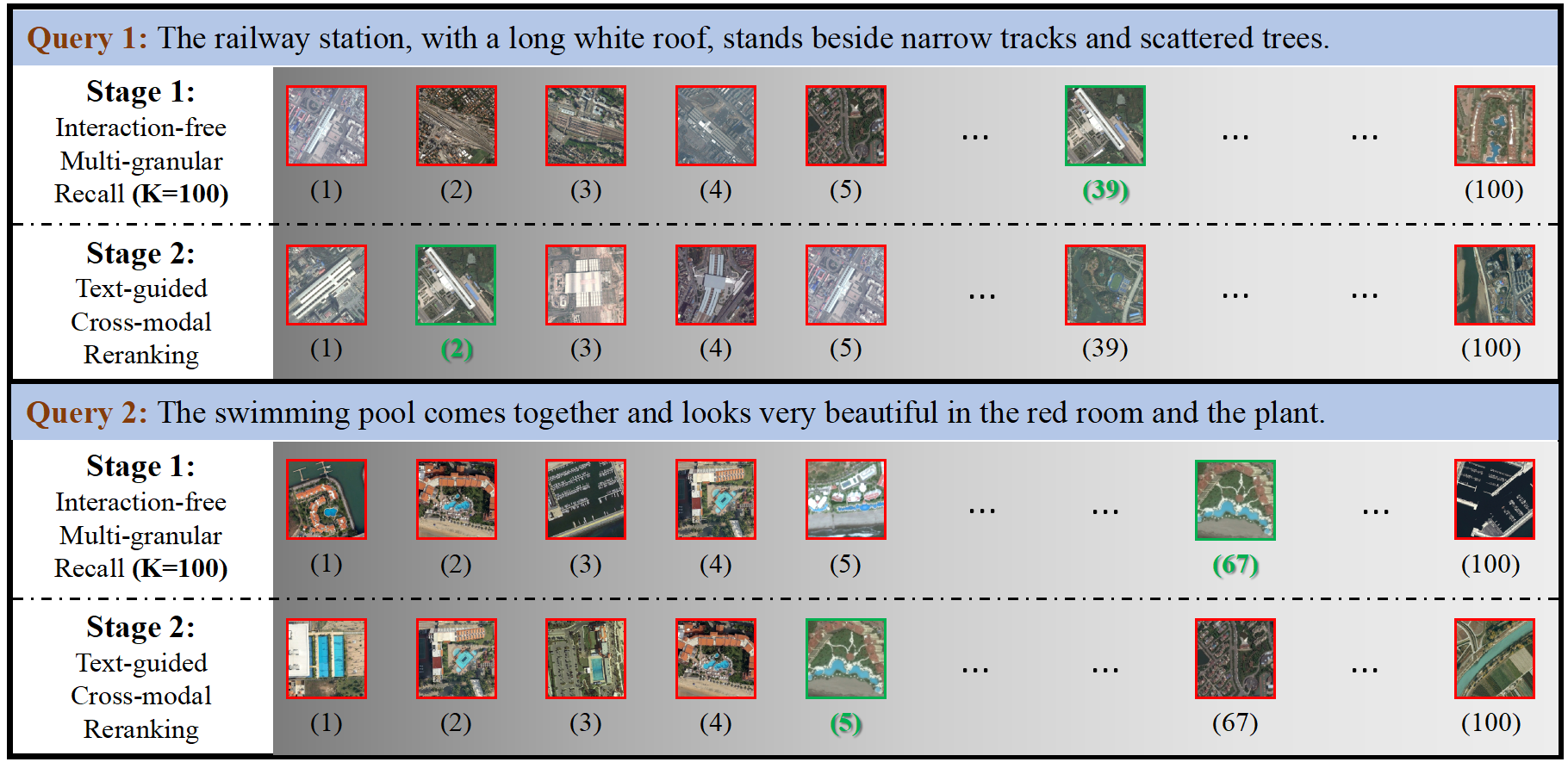}
	\caption{Qualitative examples illustrating the effectiveness of the FTF two-stage retrieval framework. In the interaction-free multi-granular recall stage, the ground-truth image is retrieved into the top-$K$ candidate set but may appear at a relatively low rank. After text-guided cross-modal reranking with BTIB, the same image is promoted to a much higher position.} 
	\Description{Two examples show ground-truth images moving from low ranks after recall to substantially higher ranks after BTIB reranking.}
	\label{fig:2stage}
\end{figure}

\textbf{Effectiveness of the FTF two-stage retrieval framework.} Fig.~\ref{fig:2stage} further illustrates the effectiveness of the proposed FTF two-stage retrieval framework. In the first stage, interaction-free multi-granular recall is able to retrieve the ground-truth image into the top-$K$ candidate set, but its ranking position may remain relatively low because this stage avoids full-gallery token-level interaction. In the second stage, text-guided cross-modal reranking with BTIB further refines semantic correspondence within the top-$K$ candidate set and significantly improves the ranking of the correct image. For example, in Query 1, the ground-truth image is ranked at position 39 after recall but is promoted to rank 2 after reranking; in Query 2, it moves from rank 67 to rank 5. These examples show that interaction-free multi-granular recall effectively preserves semantically relevant candidates, while text-guided cross-modal reranking with BTIB further improves the final ordering through fine-grained text-guided interaction.

\begin{figure}[!htbp]
	\centering
	\includegraphics[width=0.64\linewidth,height=0.42\textheight,keepaspectratio]{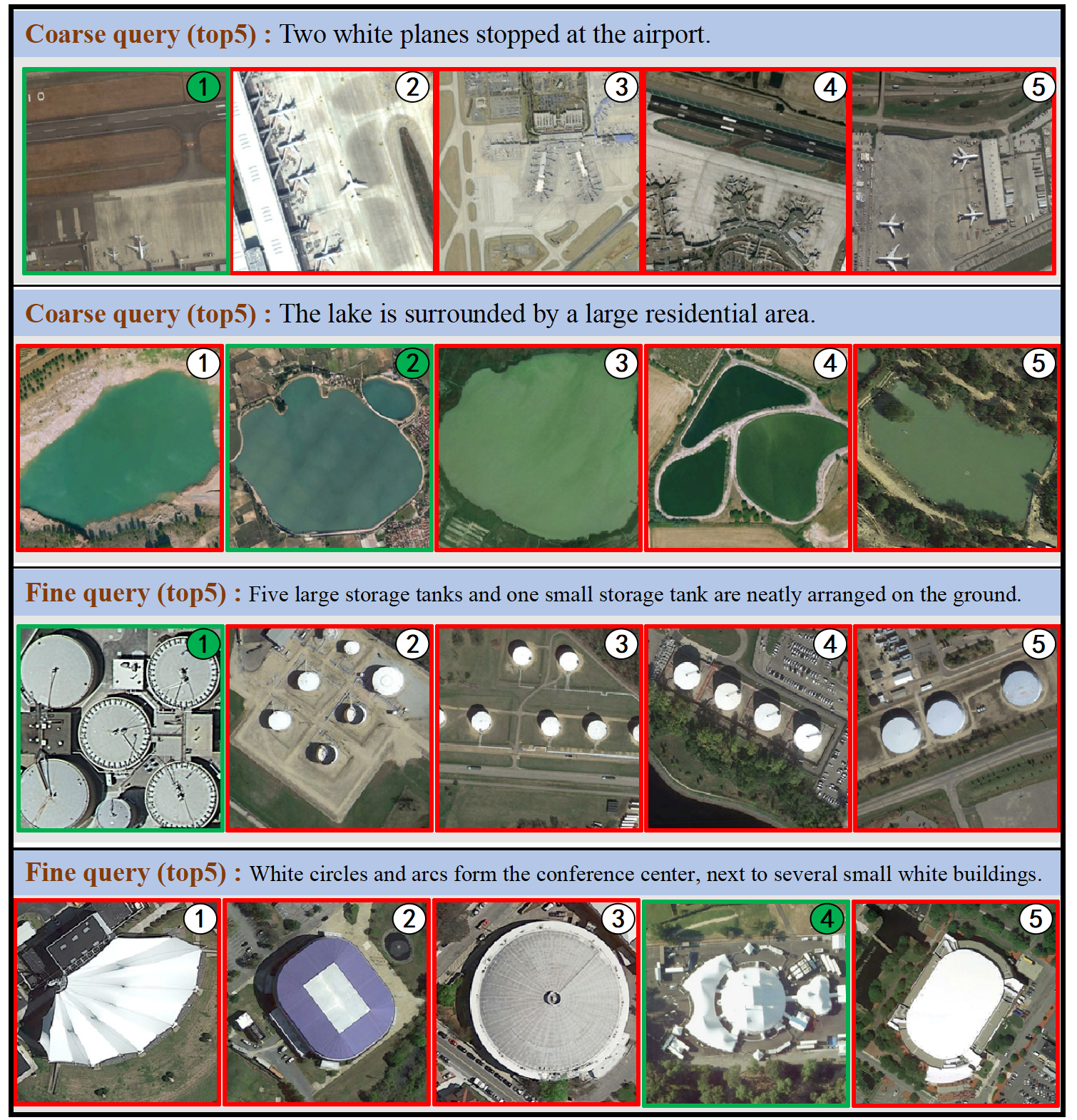}
	\caption{Qualitative examples illustrating the effectiveness of multi-granular learning. Coarse queries capture global scene structures and dominant semantic categories, while fine queries focus on local objects and structural details. Ground-truth retrievals are highlighted with green boxes.}
	\Description{Retrieval examples contrast coarse scene-level queries and fine object-level queries, showing their complementary behavior.}
	\label{fig:cfquery}
\end{figure}

\textbf{Effectiveness of multi-granular learning.} Fig.~\ref{fig:cfquery} provides qualitative examples illustrating the benefit of multi-granular learning. Coarse-grained queries capture global scene structures and dominant semantic categories, such as airports or lakes, but may retrieve scenes with similar overall layouts while overlooking subtle object differences. Fine-grained queries are more sensitive to local object arrangements and geometric details, such as the number and layout of storage tanks or the shape of buildings, although local similarities alone may be insufficient to distinguish the broader scene context. Combining coarse and fine representations therefore provides complementary retrieval cues, allowing the framework to consider both scene-level semantics and local structural details.

\FloatBarrier
\section{Conclusions}
In this work, we propose FTF, a two-stage framework that achieves competitive RS image--text retrieval accuracy with a favorable accuracy--efficiency balance. The interaction-free multi-granular recall stage selects top-$K$ candidates using pre-computed gallery embeddings, thereby avoiding costly full-gallery token-level text--image interaction. The text-guided cross-modal reranking stage then employs the parameter-free BTIB to refine semantic correspondence within the candidate set, while the proposed I$^2$M Loss jointly optimizes inter-modal alignment and intra-modal structural consistency. Extensive experiments on three public benchmarks demonstrate that FTF maintains competitive retrieval performance while substantially reducing inference cost. Moreover, the candidate size provides a flexible mechanism for balancing retrieval accuracy, latency, and throughput across different deployment requirements. These results establish FTF as an effective and scalable solution for efficient RS image--text retrieval.
\bibliographystyle{ACM-Reference-Format}
\bibliography{references}


\begin{thebibliography}{64}


\ifx \showCODEN    \undefined \def \showCODEN     #1{\unskip}     \fi
\ifx \showISBNx    \undefined \def \showISBNx     #1{\unskip}     \fi
\ifx \showISBNxiii \undefined \def \showISBNxiii  #1{\unskip}     \fi
\ifx \showISSN     \undefined \def \showISSN      #1{\unskip}     \fi
\ifx \showLCCN     \undefined \def \showLCCN      #1{\unskip}     \fi
\ifx \shownote     \undefined \def \shownote      #1{#1}          \fi
\ifx \showarticletitle \undefined \def \showarticletitle #1{#1}   \fi
\ifx \showURL      \undefined \def \showURL       {\relax}        \fi
\providecommand\bibfield[2]{#2}
\providecommand\bibinfo[2]{#2}
\providecommand\natexlab[1]{#1}
\providecommand\showeprint[2][]{arXiv:#2}

\bibitem[Asokan et~al\mbox{.}(2025)]%
        {Asokan_2025_CVPR}
\bibfield{author}{\bibinfo{person}{Mothilal Asokan}, \bibinfo{person}{Kebin
  Wu}, {and} \bibinfo{person}{Fatima Albreiki}.}
  \bibinfo{year}{2025}\natexlab{}.
\newblock \showarticletitle{{FineLIP}: Extending {CLIP}'s Reach via
  Fine-Grained Alignment with Longer Text Inputs}. In
  \bibinfo{booktitle}{\emph{Proceedings of the IEEE/CVF Conference on Computer
  Vision and Pattern Recognition (CVPR)}}. \bibinfo{pages}{14495--14504}.
\newblock
\href{https://doi.org/10.1109/CVPR52734.2025.01351}{doi:\nolinkurl{10.1109/CVPR52734.2025.01351}}


\bibitem[Astruc et~al\mbox{.}(2025)]%
        {Astruc_2025_CVPR}
\bibfield{author}{\bibinfo{person}{Guillaume Astruc}, \bibinfo{person}{Nicolas
  Gonthier}, \bibinfo{person}{Cl{\'e}ment Mallet}, {and} \bibinfo{person}{Loic
  Landrieu}.} \bibinfo{year}{2025}\natexlab{}.
\newblock \showarticletitle{AnySat: One Earth Observation Model for Many
  Resolutions, Scales, and Modalities}. In
  \bibinfo{booktitle}{\emph{Proceedings of the IEEE/CVF Conference on Computer
  Vision and Pattern Recognition (CVPR)}}. \bibinfo{pages}{19530--19540}.
\newblock
\href{https://doi.org/10.1109/CVPR52734.2025.01819}{doi:\nolinkurl{10.1109/CVPR52734.2025.01819}}


\bibitem[Chen et~al\mbox{.}(2026a)]%
        {26arxivMultiBit}
\bibfield{author}{\bibinfo{person}{Xi Chen}, \bibinfo{person}{Xu Chen},
  \bibinfo{person}{Xiangyang Jia}, \bibinfo{person}{Ting Gan},
  \bibinfo{person}{Xu Zhang}, \bibinfo{person}{Shuquan Wei}, {and}
  \bibinfo{person}{Sitong Fan}.} \bibinfo{year}{2026}\natexlab{a}.
\newblock \bibinfo{title}{Relevance-Resolution Transfer via Scale-Decomposable
  Fractional Diffusion for Multi-Length Cross-Modal Hash Retrieval}.
\newblock \bibinfo{howpublished}{arXiv:2609.34393}.
\newblock
\urldef\tempurl%
\url{https://arxiv.org/abs/2609.34393}
\showURL{%
\tempurl}


\bibitem[Chen et~al\mbox{.}(2026b)]%
        {26arxivDARAD}
\bibfield{author}{\bibinfo{person}{Xi Chen}, \bibinfo{person}{Xu Chen},
  \bibinfo{person}{Xiangyang Jia}, \bibinfo{person}{Wei Wang},
  \bibinfo{person}{Xu Zhang}, {and} \bibinfo{person}{Zhenyuan Sun}.}
  \bibinfo{year}{2026}\natexlab{b}.
\newblock \bibinfo{title}{{DARAD}: Dual Adapters and Ranking-Aware Distillation
  for Continual Remote Sensing Image-Text Retrieval}.
\newblock \bibinfo{howpublished}{arXiv:2608.06059}.
\newblock
\urldef\tempurl%
\url{https://arxiv.org/abs/2608.06059}
\showURL{%
\tempurl}


\bibitem[Chen et~al\mbox{.}(2026c)]%
        {26tgrsGraVLM}
\bibfield{author}{\bibinfo{person}{Xi Chen}, \bibinfo{person}{Xu Chen},
  \bibinfo{person}{Wei Wang}, \bibinfo{person}{Xiangyang Jia},
  \bibinfo{person}{Gongping Yu}, {and} \bibinfo{person}{Feng Li}.}
  \bibinfo{year}{2026}\natexlab{c}.
\newblock \showarticletitle{{GraVLM}: A Hierarchical Graph-Aligned
  Vision-Language Model for Cross-Modal Retrieval in Remote Sensing}.
\newblock \bibinfo{journal}{\emph{IEEE Transactions on Geoscience and Remote
  Sensing}}  \bibinfo{volume}{64} (\bibinfo{year}{2026}),
  \bibinfo{pages}{1--15}.
\newblock
\href{https://doi.org/10.1109/TGRS.2026.3671643}{doi:\nolinkurl{10.1109/TGRS.2026.3671643}}


\bibitem[Chen et~al\mbox{.}(2025)]%
        {25tgrsCLGSA}
\bibfield{author}{\bibinfo{person}{Xiumei Chen}, \bibinfo{person}{Xiangtao
  Zheng}, {and} \bibinfo{person}{Xiaoqiang Lu}.}
  \bibinfo{year}{2025}\natexlab{}.
\newblock \showarticletitle{Context-Aware Local-Global Semantic Alignment for
  Remote Sensing Image-Text Retrieval}.
\newblock \bibinfo{journal}{\emph{IEEE Transactions on Geoscience and Remote
  Sensing}}  \bibinfo{volume}{63} (\bibinfo{year}{2025}),
  \bibinfo{pages}{1--12}.
\newblock
\href{https://doi.org/10.1109/TGRS.2025.3552304}{doi:\nolinkurl{10.1109/TGRS.2025.3552304}}


\bibitem[Cheng et~al\mbox{.}(2024)]%
        {24prModalInteraction}
\bibfield{author}{\bibinfo{person}{Hang Cheng}, \bibinfo{person}{Hehui Ye},
  \bibinfo{person}{Xiaofei Zhou}, \bibinfo{person}{Ximeng Liu},
  \bibinfo{person}{Fei Chen}, {and} \bibinfo{person}{Meiqing Wang}.}
  \bibinfo{year}{2024}\natexlab{}.
\newblock \showarticletitle{Vision-Language Pre-Training via Modal
  Interaction}.
\newblock \bibinfo{journal}{\emph{Pattern Recognition}}  \bibinfo{volume}{156}
  (\bibinfo{year}{2024}), \bibinfo{pages}{110809}.
\newblock
\href{https://doi.org/10.1016/j.patcog.2024.110809}{doi:\nolinkurl{10.1016/j.patcog.2024.110809}}


\bibitem[Choi et~al\mbox{.}(2025)]%
        {25cvprGOAL}
\bibfield{author}{\bibinfo{person}{Hyungyu Choi}, \bibinfo{person}{Young~Kyun
  Jang}, {and} \bibinfo{person}{Chanho Eom}.} \bibinfo{year}{2025}\natexlab{}.
\newblock \showarticletitle{{GOAL}: Global-Local Object Alignment Learning}. In
  \bibinfo{booktitle}{\emph{Proceedings of the IEEE/CVF Conference on Computer
  Vision and Pattern Recognition (CVPR)}}. \bibinfo{pages}{4070--4079}.
\newblock
\href{https://doi.org/10.1109/CVPR52734.2025.00385}{doi:\nolinkurl{10.1109/CVPR52734.2025.00385}}


\bibitem[Grattafiori et~al\mbox{.}(2024)]%
        {24arxivLlama3}
\bibfield{author}{\bibinfo{person}{Aaron Grattafiori},
  \bibinfo{person}{Abhimanyu Dubey}, {et~al\mbox{.}}}
  \bibinfo{year}{2024}\natexlab{}.
\newblock \bibinfo{title}{The {Llama} 3 Herd of Models}.
\newblock \bibinfo{howpublished}{arXiv:2407.21783}.
\newblock
\urldef\tempurl%
\url{https://arxiv.org/abs/2407.21783}
\showURL{%
\tempurl}


\bibitem[He et~al\mbox{.}(2024)]%
        {24tgrsVGSGN}
\bibfield{author}{\bibinfo{person}{Yangpeng He}, \bibinfo{person}{Xin Xu},
  \bibinfo{person}{Hongjia Chen}, \bibinfo{person}{Jinwen Li}, {and}
  \bibinfo{person}{Fangling Pu}.} \bibinfo{year}{2024}\natexlab{}.
\newblock \showarticletitle{Visual Global-Salient-Guided Network for Remote
  Sensing Image-Text Retrieval}.
\newblock \bibinfo{journal}{\emph{IEEE Transactions on Geoscience and Remote
  Sensing}}  \bibinfo{volume}{62} (\bibinfo{year}{2024}),
  \bibinfo{pages}{1--14}.
\newblock
\href{https://doi.org/10.1109/TGRS.2024.3466389}{doi:\nolinkurl{10.1109/TGRS.2024.3466389}}


\bibitem[Huang et~al\mbox{.}(2025)]%
        {huang2025noiserobust}
\bibfield{author}{\bibinfo{person}{Zhenyu Huang}, \bibinfo{person}{Mouxing
  Yang}, \bibinfo{person}{Xinyan Xiao}, \bibinfo{person}{Peng Hu}, {and}
  \bibinfo{person}{Xi Peng}.} \bibinfo{year}{2025}\natexlab{}.
\newblock \showarticletitle{Noise-Robust Vision-Language Pre-Training with
  Positive-Negative Learning}.
\newblock \bibinfo{journal}{\emph{IEEE Transactions on Pattern Analysis and
  Machine Intelligence}} \bibinfo{volume}{47}, \bibinfo{number}{1}
  (\bibinfo{year}{2025}), \bibinfo{pages}{338--350}.
\newblock
\href{https://doi.org/10.1109/TPAMI.2024.3462996}{doi:\nolinkurl{10.1109/TPAMI.2024.3462996}}


\bibitem[Ji et~al\mbox{.}(2023)]%
        {23tgrsKAMCL}
\bibfield{author}{\bibinfo{person}{Zhong Ji}, \bibinfo{person}{Changxu Meng},
  \bibinfo{person}{Yan Zhang}, \bibinfo{person}{Yanwei Pang}, {and}
  \bibinfo{person}{Xuelong Li}.} \bibinfo{year}{2023}\natexlab{}.
\newblock \showarticletitle{Knowledge-Aided Momentum Contrastive Learning for
  Remote-Sensing Image Text Retrieval}.
\newblock \bibinfo{journal}{\emph{IEEE Transactions on Geoscience and Remote
  Sensing}}  \bibinfo{volume}{61} (\bibinfo{year}{2023}),
  \bibinfo{pages}{1--13}.
\newblock
\href{https://doi.org/10.1109/TGRS.2023.3332317}{doi:\nolinkurl{10.1109/TGRS.2023.3332317}}


\bibitem[Ji et~al\mbox{.}(2024)]%
        {24acmmmEBAKER}
\bibfield{author}{\bibinfo{person}{Zhong Ji}, \bibinfo{person}{Changxu Meng},
  \bibinfo{person}{Yan Zhang}, \bibinfo{person}{Haoran Wang},
  \bibinfo{person}{Yanwei Pang}, {and} \bibinfo{person}{Jungong Han}.}
  \bibinfo{year}{2024}\natexlab{}.
\newblock \showarticletitle{Eliminate Before Align: A Remote Sensing Image-Text
  Retrieval Framework with Keyword Explicit Reasoning}. In
  \bibinfo{booktitle}{\emph{Proceedings of the 32nd ACM International
  Conference on Multimedia}}. \bibinfo{pages}{1662--1671}.
\newblock
\href{https://doi.org/10.1145/3664647.3681270}{doi:\nolinkurl{10.1145/3664647.3681270}}


\bibitem[Jiang et~al\mbox{.}(2025)]%
        {25patrecCMA}
\bibfield{author}{\bibinfo{person}{Haojun Jiang}, \bibinfo{person}{Jianke
  Zhang}, \bibinfo{person}{Rui Huang}, \bibinfo{person}{Chunjiang Ge},
  \bibinfo{person}{Zanlin Ni}, \bibinfo{person}{Shiji Song}, {and}
  \bibinfo{person}{Gao Huang}.} \bibinfo{year}{2025}\natexlab{}.
\newblock \showarticletitle{Cross-Modal Adapter for Vision-Language Retrieval}.
\newblock \bibinfo{journal}{\emph{Pattern Recognition}}  \bibinfo{volume}{159}
  (\bibinfo{year}{2025}), \bibinfo{pages}{111144}.
\newblock
\href{https://doi.org/10.1016/j.patcog.2024.111144}{doi:\nolinkurl{10.1016/j.patcog.2024.111144}}


\bibitem[Kim et~al\mbox{.}(2024)]%
        {24eccvOTSeg}
\bibfield{author}{\bibinfo{person}{Kwanyoung Kim}, \bibinfo{person}{Yujin Oh},
  {and} \bibinfo{person}{Jong~Chul Ye}.} \bibinfo{year}{2024}\natexlab{}.
\newblock \showarticletitle{{OTSeg}: Multi-Prompt Sinkhorn Attention for
  Zero-Shot Semantic Segmentation}. In \bibinfo{booktitle}{\emph{Computer
  Vision -- ECCV 2024}}. \bibinfo{pages}{200--217}.
\newblock
\href{https://doi.org/10.1007/978-3-031-72980-5_12}{doi:\nolinkurl{10.1007/978-3-031-72980-5_12}}


\bibitem[Lee et~al\mbox{.}(2018)]%
        {2018eccvSCAN}
\bibfield{author}{\bibinfo{person}{Kuang-Huei Lee}, \bibinfo{person}{Xi Chen},
  \bibinfo{person}{Gang Hua}, \bibinfo{person}{Houdong Hu}, {and}
  \bibinfo{person}{Xiaodong He}.} \bibinfo{year}{2018}\natexlab{}.
\newblock \showarticletitle{Stacked Cross Attention for Image-Text Matching}.
  In \bibinfo{booktitle}{\emph{Computer Vision -- ECCV 2018}}.
  \bibinfo{pages}{212--228}.
\newblock
\href{https://doi.org/10.1007/978-3-030-01225-0_13}{doi:\nolinkurl{10.1007/978-3-030-01225-0_13}}


\bibitem[Li and Li(2026)]%
        {26tmmCIGNN}
\bibfield{author}{\bibinfo{person}{Bo Li} {and} \bibinfo{person}{Zhixin Li}.}
  \bibinfo{year}{2026}\natexlab{}.
\newblock \showarticletitle{Causality-Inspired Graph Neural Networks for
  Cross-Modal Retrieval}.
\newblock \bibinfo{journal}{\emph{IEEE Transactions on Multimedia}}
  \bibinfo{volume}{28} (\bibinfo{year}{2026}), \bibinfo{pages}{2497--2509}.
\newblock
\href{https://doi.org/10.1109/TMM.2026.3651028}{doi:\nolinkurl{10.1109/TMM.2026.3651028}}


\bibitem[Li et~al\mbox{.}(2023)]%
        {23icmlBLIP2}
\bibfield{author}{\bibinfo{person}{Junnan Li}, \bibinfo{person}{Dongxu Li},
  \bibinfo{person}{Silvio Savarese}, {and} \bibinfo{person}{Steven C.~H. Hoi}.}
  \bibinfo{year}{2023}\natexlab{}.
\newblock \showarticletitle{{BLIP-2}: Bootstrapping Language-Image Pre-Training
  with Frozen Image Encoders and Large Language Models}. In
  \bibinfo{booktitle}{\emph{Proceedings of the 40th International Conference on
  Machine Learning}} \emph{(\bibinfo{series}{Proceedings of Machine Learning
  Research}, Vol.~\bibinfo{volume}{202})}. \bibinfo{publisher}{PMLR},
  \bibinfo{pages}{19730--19742}.
\newblock
\urldef\tempurl%
\url{https://proceedings.mlr.press/v202/li23q.html}
\showURL{%
\tempurl}


\bibitem[Li et~al\mbox{.}(2022)]%
        {22icmlBLIP}
\bibfield{author}{\bibinfo{person}{Junnan Li}, \bibinfo{person}{Dongxu Li},
  \bibinfo{person}{Caiming Xiong}, {and} \bibinfo{person}{Steven C.~H. Hoi}.}
  \bibinfo{year}{2022}\natexlab{}.
\newblock \showarticletitle{{BLIP}: Bootstrapping Language-Image Pre-Training
  for Unified Vision-Language Understanding and Generation}. In
  \bibinfo{booktitle}{\emph{Proceedings of the 39th International Conference on
  Machine Learning}} \emph{(\bibinfo{series}{Proceedings of Machine Learning
  Research}, Vol.~\bibinfo{volume}{162})}. \bibinfo{publisher}{PMLR},
  \bibinfo{pages}{12888--12900}.
\newblock
\urldef\tempurl%
\url{https://proceedings.mlr.press/v162/li22n.html}
\showURL{%
\tempurl}


\bibitem[Li et~al\mbox{.}(2021)]%
        {21neuripsALBEF}
\bibfield{author}{\bibinfo{person}{Junnan Li}, \bibinfo{person}{Ramprasaath~R.
  Selvaraju}, \bibinfo{person}{Akhilesh~Deepak Gotmare},
  \bibinfo{person}{Shafiq Joty}, \bibinfo{person}{Caiming Xiong}, {and}
  \bibinfo{person}{Steven C.~H. Hoi}.} \bibinfo{year}{2021}\natexlab{}.
\newblock \showarticletitle{Align before Fuse: Vision and Language
  Representation Learning with Momentum Distillation}. In
  \bibinfo{booktitle}{\emph{Advances in Neural Information Processing
  Systems}}, Vol.~\bibinfo{volume}{34}. \bibinfo{pages}{9694--9705}.
\newblock


\bibitem[Li et~al\mbox{.}(2025a)]%
        {25acmmmEAPA}
\bibfield{author}{\bibinfo{person}{Shuoshuo Li}, \bibinfo{person}{Shuli Cheng},
  {and} \bibinfo{person}{Liejun Wang}.} \bibinfo{year}{2025}\natexlab{a}.
\newblock \showarticletitle{Entity-Level Alignment with Prompt-Guided Adapter
  for Remote Sensing Image-Text Retrieval}. In
  \bibinfo{booktitle}{\emph{Proceedings of the 33rd ACM International
  Conference on Multimedia}}. \bibinfo{pages}{8224--8233}.
\newblock
\href{https://doi.org/10.1145/3746027.3755374}{doi:\nolinkurl{10.1145/3746027.3755374}}


\bibitem[Li et~al\mbox{.}(2025b)]%
        {25airCSAN}
\bibfield{author}{\bibinfo{person}{Zhuoyi Li}, \bibinfo{person}{Huibin Lu},
  \bibinfo{person}{Hao Fu}, \bibinfo{person}{Fanzhen Meng}, {and}
  \bibinfo{person}{Guanghua Gu}.} \bibinfo{year}{2025}\natexlab{b}.
\newblock \showarticletitle{{CSAN}: Cross-Coupled Semantic Adversarial Network
  for Cross-Modal Retrieval}.
\newblock \bibinfo{journal}{\emph{Artificial Intelligence Review}}
  \bibinfo{volume}{58}, \bibinfo{number}{5} (\bibinfo{year}{2025}),
  \bibinfo{pages}{151}.
\newblock
\href{https://doi.org/10.1007/s10462-025-11152-7}{doi:\nolinkurl{10.1007/s10462-025-11152-7}}


\bibitem[Li et~al\mbox{.}(2025c)]%
        {li2025lhrs}
\bibfield{author}{\bibinfo{person}{Zhenshi Li}, \bibinfo{person}{Dilxat
  Muhtar}, \bibinfo{person}{Feng Gu}, \bibinfo{person}{Yanglangxing He},
  \bibinfo{person}{Xueliang Zhang}, \bibinfo{person}{Pengfeng Xiao},
  \bibinfo{person}{Guangjun He}, {and} \bibinfo{person}{Xiaoxiang Zhu}.}
  \bibinfo{year}{2025}\natexlab{c}.
\newblock \showarticletitle{{LHRS-Bot-Nova}: Improved Multimodal Large Language
  Model for Remote Sensing Vision-Language Interpretation}.
\newblock \bibinfo{journal}{\emph{ISPRS Journal of Photogrammetry and Remote
  Sensing}}  \bibinfo{volume}{227} (\bibinfo{year}{2025}),
  \bibinfo{pages}{539--550}.
\newblock
\href{https://doi.org/10.1016/j.isprsjprs.2025.06.003}{doi:\nolinkurl{10.1016/j.isprsjprs.2025.06.003}}


\bibitem[Liu et~al\mbox{.}(2024a)]%
        {24tgrsremoteclip}
\bibfield{author}{\bibinfo{person}{Fan Liu}, \bibinfo{person}{Delong Chen},
  \bibinfo{person}{Zhangqingyun Guan}, \bibinfo{person}{Xiaocong Zhou},
  \bibinfo{person}{Jiale Zhu}, \bibinfo{person}{Qiaolin Ye},
  \bibinfo{person}{Liyong Fu}, {and} \bibinfo{person}{Jun Zhou}.}
  \bibinfo{year}{2024}\natexlab{a}.
\newblock \showarticletitle{{RemoteCLIP}: A Vision-Language Foundation Model
  for Remote Sensing}.
\newblock \bibinfo{journal}{\emph{IEEE Transactions on Geoscience and Remote
  Sensing}}  \bibinfo{volume}{62} (\bibinfo{year}{2024}),
  \bibinfo{pages}{1--16}.
\newblock
\href{https://doi.org/10.1109/TGRS.2024.3390838}{doi:\nolinkurl{10.1109/TGRS.2024.3390838}}


\bibitem[Liu et~al\mbox{.}(2024b)]%
        {24asocCCCL}
\bibfield{author}{\bibinfo{person}{Zheng Liu}, \bibinfo{person}{Xinlei Pei},
  \bibinfo{person}{Shanshan Gao}, \bibinfo{person}{Changhao Li},
  \bibinfo{person}{Jingyao Wang}, {and} \bibinfo{person}{Junhao Xu}.}
  \bibinfo{year}{2024}\natexlab{b}.
\newblock \showarticletitle{Perceive, Reason, and Align: Context-Guided
  Cross-Modal Correlation Learning for Image-Text Retrieval}.
\newblock \bibinfo{journal}{\emph{Applied Soft Computing}}
  \bibinfo{volume}{154} (\bibinfo{year}{2024}), \bibinfo{pages}{111395}.
\newblock
\href{https://doi.org/10.1016/j.asoc.2024.111395}{doi:\nolinkurl{10.1016/j.asoc.2024.111395}}


\bibitem[Lu et~al\mbox{.}(2024)]%
        {24arxivDeepSeekVL}
\bibfield{author}{\bibinfo{person}{Haoyu Lu}, \bibinfo{person}{Wen Liu},
  \bibinfo{person}{Bo Zhang}, \bibinfo{person}{Bingxuan Wang},
  \bibinfo{person}{Kai Dong}, \bibinfo{person}{Bo Liu},
  \bibinfo{person}{Jingxiang Sun}, \bibinfo{person}{Tongzheng Ren},
  \bibinfo{person}{Zhuoshu Li}, \bibinfo{person}{Hao Yang},
  \bibinfo{person}{Yaofeng Sun}, \bibinfo{person}{Chengqi Deng},
  \bibinfo{person}{Hanwei Xu}, \bibinfo{person}{Zhenda Xie}, {and}
  \bibinfo{person}{Chong Ruan}.} \bibinfo{year}{2024}\natexlab{}.
\newblock \bibinfo{title}{{DeepSeek-VL}: Towards Real-World Vision-Language
  Understanding}.
\newblock \bibinfo{howpublished}{arXiv:2403.05525}.
\newblock
\urldef\tempurl%
\url{https://arxiv.org/abs/2403.05525}
\showURL{%
\tempurl}


\bibitem[Lu et~al\mbox{.}(2018)]%
        {18tgrsRSICD}
\bibfield{author}{\bibinfo{person}{Xiaoqiang Lu}, \bibinfo{person}{Binqiang
  Wang}, \bibinfo{person}{Xiangtao Zheng}, {and} \bibinfo{person}{Xuelong Li}.}
  \bibinfo{year}{2018}\natexlab{}.
\newblock \showarticletitle{Exploring Models and Data for Remote Sensing Image
  Caption Generation}.
\newblock \bibinfo{journal}{\emph{IEEE Transactions on Geoscience and Remote
  Sensing}} \bibinfo{volume}{56}, \bibinfo{number}{4} (\bibinfo{year}{2018}),
  \bibinfo{pages}{2183--2195}.
\newblock
\href{https://doi.org/10.1109/TGRS.2017.2776321}{doi:\nolinkurl{10.1109/TGRS.2017.2776321}}


\bibitem[Luo et~al\mbox{.}(2022)]%
        {22neurocomputingCLIP4Clip}
\bibfield{author}{\bibinfo{person}{Huaishao Luo}, \bibinfo{person}{Lei Ji},
  \bibinfo{person}{Ming Zhong}, \bibinfo{person}{Yang Chen},
  \bibinfo{person}{Wen Lei}, \bibinfo{person}{Nan Duan}, {and}
  \bibinfo{person}{Tianrui Li}.} \bibinfo{year}{2022}\natexlab{}.
\newblock \showarticletitle{{CLIP4Clip}: An Empirical Study of {CLIP} for
  End-to-End Video Clip Retrieval and Captioning}.
\newblock \bibinfo{journal}{\emph{Neurocomputing}}  \bibinfo{volume}{508}
  (\bibinfo{year}{2022}), \bibinfo{pages}{293--304}.
\newblock
\href{https://doi.org/10.1016/j.neucom.2022.07.028}{doi:\nolinkurl{10.1016/j.neucom.2022.07.028}}


\bibitem[Ma et~al\mbox{.}(2025)]%
        {ma2025multimodal}
\bibfield{author}{\bibinfo{person}{Zehong Ma}, \bibinfo{person}{Hao Chen},
  \bibinfo{person}{Wei Zeng}, \bibinfo{person}{Limin Su}, {and}
  \bibinfo{person}{Shiliang Zhang}.} \bibinfo{year}{2025}\natexlab{}.
\newblock \showarticletitle{Multi-Modal Reference Learning for Fine-Grained
  Text-to-Image Retrieval}.
\newblock \bibinfo{journal}{\emph{IEEE Transactions on Multimedia}}
  \bibinfo{volume}{27} (\bibinfo{year}{2025}), \bibinfo{pages}{5009--5022}.
\newblock
\href{https://doi.org/10.1109/TMM.2025.3543066}{doi:\nolinkurl{10.1109/TMM.2025.3543066}}


\bibitem[Mall et~al\mbox{.}(2024)]%
        {24iclrGRAFT}
\bibfield{author}{\bibinfo{person}{Utkarsh Mall}, \bibinfo{person}{Cheng~Perng
  Phoo}, \bibinfo{person}{Meilin~Kelsey Liu}, \bibinfo{person}{Carl Vondrick},
  \bibinfo{person}{Bharath Hariharan}, {and} \bibinfo{person}{Kavita Bala}.}
  \bibinfo{year}{2024}\natexlab{}.
\newblock \showarticletitle{Remote Sensing Vision-Language Foundation Models
  Without Annotations via Ground Remote Alignment}. In
  \bibinfo{booktitle}{\emph{The Twelfth International Conference on Learning
  Representations}}.
\newblock
\urldef\tempurl%
\url{https://openreview.net/forum?id=w9tc699w3Z}
\showURL{%
\tempurl}


\bibitem[Naeem et~al\mbox{.}(2024)]%
        {24eccvSILC}
\bibfield{author}{\bibinfo{person}{Muhammad~Ferjad Naeem},
  \bibinfo{person}{Yongqin Xian}, \bibinfo{person}{Xiaohua Zhai},
  \bibinfo{person}{Lukas Hoyer}, \bibinfo{person}{Luc Van~Gool}, {and}
  \bibinfo{person}{Federico Tombari}.} \bibinfo{year}{2024}\natexlab{}.
\newblock \showarticletitle{{SILC}: Improving Vision Language Pretraining with
  Self-Distillation}. In \bibinfo{booktitle}{\emph{Computer Vision -- ECCV
  2024}}. \bibinfo{pages}{38--55}.
\newblock
\href{https://doi.org/10.1007/978-3-031-72664-4_3}{doi:\nolinkurl{10.1007/978-3-031-72664-4_3}}


\bibitem[Ouyang et~al\mbox{.}(2025)]%
        {25tgrsSIPN}
\bibfield{author}{\bibinfo{person}{Pengxiang Ouyang}, \bibinfo{person}{Qing
  Ma}, {and} \bibinfo{person}{Cong Bai}.} \bibinfo{year}{2025}\natexlab{}.
\newblock \showarticletitle{Sparse Information Perception Network for Remote
  Sensing Cross-Modal Retrieval}.
\newblock \bibinfo{journal}{\emph{IEEE Transactions on Geoscience and Remote
  Sensing}}  \bibinfo{volume}{63} (\bibinfo{year}{2025}),
  \bibinfo{pages}{1--15}.
\newblock
\href{https://doi.org/10.1109/TGRS.2025.3597055}{doi:\nolinkurl{10.1109/TGRS.2025.3597055}}


\bibitem[Pan et~al\mbox{.}(2023)]%
        {23acmmmPIR}
\bibfield{author}{\bibinfo{person}{Jiancheng Pan}, \bibinfo{person}{Qing Ma},
  {and} \bibinfo{person}{Cong Bai}.} \bibinfo{year}{2023}\natexlab{}.
\newblock \showarticletitle{A Prior Instruction Representation Framework for
  Remote Sensing Image-Text Retrieval}. In
  \bibinfo{booktitle}{\emph{Proceedings of the 31st ACM International
  Conference on Multimedia}}. \bibinfo{pages}{611--620}.
\newblock
\href{https://doi.org/10.1145/3581783.3612374}{doi:\nolinkurl{10.1145/3581783.3612374}}


\bibitem[Pan et~al\mbox{.}(2025)]%
        {25tommReAL}
\bibfield{author}{\bibinfo{person}{Renjie Pan}, \bibinfo{person}{Hua Yang},
  {and} \bibinfo{person}{Xiangyu Zhao}.} \bibinfo{year}{2025}\natexlab{}.
\newblock \showarticletitle{{ReAL}: Improving Image-Text Retrieval with
  Authentic Negative Repository Learning}.
\newblock \bibinfo{journal}{\emph{ACM Transactions on Multimedia Computing,
  Communications, and Applications}} \bibinfo{volume}{21},
  \bibinfo{number}{10}, Article \bibinfo{articleno}{279}
  (\bibinfo{year}{2025}), \bibinfo{numpages}{22}~pages.
\newblock
\href{https://doi.org/10.1145/3729172}{doi:\nolinkurl{10.1145/3729172}}


\bibitem[Qu et~al\mbox{.}(2016)]%
        {16citsUCMCaptions}
\bibfield{author}{\bibinfo{person}{Bo Qu}, \bibinfo{person}{Xuelong Li},
  \bibinfo{person}{Dacheng Tao}, {and} \bibinfo{person}{Xiaoqiang Lu}.}
  \bibinfo{year}{2016}\natexlab{}.
\newblock \showarticletitle{Deep Semantic Understanding of High Resolution
  Remote Sensing Image}. In \bibinfo{booktitle}{\emph{2016 International
  Conference on Computer, Information and Telecommunication Systems (CITS)}}.
  \bibinfo{pages}{1--5}.
\newblock
\href{https://doi.org/10.1109/CITS.2016.7546397}{doi:\nolinkurl{10.1109/CITS.2016.7546397}}


\bibitem[Radford et~al\mbox{.}(2021)]%
        {21icmlCLIPBASE}
\bibfield{author}{\bibinfo{person}{Alec Radford}, \bibinfo{person}{Jong~Wook
  Kim}, \bibinfo{person}{Chris Hallacy}, \bibinfo{person}{Aditya Ramesh},
  \bibinfo{person}{Gabriel Goh}, \bibinfo{person}{Sandhini Agarwal},
  \bibinfo{person}{Girish Sastry}, \bibinfo{person}{Amanda Askell},
  \bibinfo{person}{Pamela Mishkin}, \bibinfo{person}{Jack Clark},
  \bibinfo{person}{Gretchen Krueger}, {and} \bibinfo{person}{Ilya Sutskever}.}
  \bibinfo{year}{2021}\natexlab{}.
\newblock \showarticletitle{Learning Transferable Visual Models from Natural
  Language Supervision}. In \bibinfo{booktitle}{\emph{Proceedings of the 38th
  International Conference on Machine Learning}}
  \emph{(\bibinfo{series}{Proceedings of Machine Learning Research},
  Vol.~\bibinfo{volume}{139})}. \bibinfo{publisher}{PMLR},
  \bibinfo{pages}{8748--8763}.
\newblock


\bibitem[Sha et~al\mbox{.}(2025)]%
        {25tommEEDR}
\bibfield{author}{\bibinfo{person}{Yuchen Sha}, \bibinfo{person}{Yujian Feng},
  \bibinfo{person}{Jing Sun}, \bibinfo{person}{Yichi Jin},
  \bibinfo{person}{Shuai You}, \bibinfo{person}{Lingli Wan},
  \bibinfo{person}{Yimu Ji}, {and} \bibinfo{person}{Shangdong Liu}.}
  \bibinfo{year}{2025}\natexlab{}.
\newblock \showarticletitle{Enhancing Embedding Diversity and Robustness for
  Image-Text Retrieval in Remote Sensing}.
\newblock \bibinfo{journal}{\emph{ACM Transactions on Multimedia Computing,
  Communications, and Applications}}, Article \bibinfo{articleno}{3773281}
  (\bibinfo{year}{2025}), \bibinfo{numpages}{21}~pages.
\newblock
\href{https://doi.org/10.1145/3773281}{doi:\nolinkurl{10.1145/3773281}}


\bibitem[Tang et~al\mbox{.}(2017)]%
        {tsr2017}
\bibfield{author}{\bibinfo{person}{Xu Tang}, \bibinfo{person}{Licheng Jiao},
  \bibinfo{person}{William~J. Emery}, \bibinfo{person}{Fang Liu}, {and}
  \bibinfo{person}{Dan Zhang}.} \bibinfo{year}{2017}\natexlab{}.
\newblock \showarticletitle{Two-Stage Reranking for Remote Sensing Image
  Retrieval}.
\newblock \bibinfo{journal}{\emph{IEEE Transactions on Geoscience and Remote
  Sensing}} \bibinfo{volume}{55}, \bibinfo{number}{10} (\bibinfo{year}{2017}),
  \bibinfo{pages}{5798--5817}.
\newblock
\href{https://doi.org/10.1109/TGRS.2017.2714676}{doi:\nolinkurl{10.1109/TGRS.2017.2714676}}


\bibitem[Tu et~al\mbox{.}(2025)]%
        {25tpamiGLSCL}
\bibfield{author}{\bibinfo{person}{Rong-Cheng Tu}, \bibinfo{person}{Yatai Ji},
  \bibinfo{person}{Jie Jiang}, \bibinfo{person}{Weijie Kong},
  \bibinfo{person}{Chengfei Cai}, \bibinfo{person}{Wenzhe Zhao},
  \bibinfo{person}{Hongfa Wang}, \bibinfo{person}{Yujiu Yang}, {and}
  \bibinfo{person}{Wei Liu}.} \bibinfo{year}{2025}\natexlab{}.
\newblock \showarticletitle{Global and Local Semantic Completion Learning for
  Vision-Language Pre-Training}.
\newblock \bibinfo{journal}{\emph{IEEE Transactions on Pattern Analysis and
  Machine Intelligence}} \bibinfo{volume}{47}, \bibinfo{number}{12}
  (\bibinfo{year}{2025}), \bibinfo{pages}{11065--11079}.
\newblock
\href{https://doi.org/10.1109/TPAMI.2025.3596394}{doi:\nolinkurl{10.1109/TPAMI.2025.3596394}}


\bibitem[Wang et~al\mbox{.}(2026a)]%
        {25neunetSFE}
\bibfield{author}{\bibinfo{person}{Jingyao Wang}, \bibinfo{person}{Zheng Liu},
  \bibinfo{person}{Shanshan Gao}, \bibinfo{person}{Junhao Xu}, {and}
  \bibinfo{person}{Changhao Li}.} \bibinfo{year}{2026}\natexlab{a}.
\newblock \showarticletitle{From External to Internal: Step-Wise Feature
  Enhancement Network for Image-Text Retrieval}.
\newblock \bibinfo{journal}{\emph{Neural Networks}}  \bibinfo{volume}{193}
  (\bibinfo{year}{2026}), \bibinfo{pages}{108072}.
\newblock
\href{https://doi.org/10.1016/j.neunet.2025.108072}{doi:\nolinkurl{10.1016/j.neunet.2025.108072}}


\bibitem[Wang et~al\mbox{.}(2026b)]%
        {26tmmCaption2CLIP}
\bibfield{author}{\bibinfo{person}{Teng Wang}, \bibinfo{person}{Ruize Wang},
  \bibinfo{person}{Lingquan Meng}, {and} \bibinfo{person}{Changyin Sun}.}
  \bibinfo{year}{2026}\natexlab{b}.
\newblock \showarticletitle{Multimodal Large Language Models Assisted
  Hierarchical Image-Caption Fusion for Remote Sensing Image-Text Retrieval}.
\newblock \bibinfo{journal}{\emph{IEEE Transactions on Multimedia}}
  \bibinfo{volume}{28} (\bibinfo{year}{2026}), \bibinfo{pages}{7621--7634}.
\newblock
\href{https://doi.org/10.1109/TMM.2026.3675936}{doi:\nolinkurl{10.1109/TMM.2026.3675936}}


\bibitem[Wang et~al\mbox{.}(2019)]%
        {2019iccvCAMP}
\bibfield{author}{\bibinfo{person}{Zihao Wang}, \bibinfo{person}{Xihui Liu},
  \bibinfo{person}{Hongsheng Li}, \bibinfo{person}{Lu Sheng},
  \bibinfo{person}{Junjie Yan}, \bibinfo{person}{Xiaogang Wang}, {and}
  \bibinfo{person}{Jing Shao}.} \bibinfo{year}{2019}\natexlab{}.
\newblock \showarticletitle{{CAMP}: Cross-Modal Adaptive Message Passing for
  Text-Image Retrieval}. In \bibinfo{booktitle}{\emph{Proceedings of the
  IEEE/CVF International Conference on Computer Vision (ICCV)}}.
  \bibinfo{pages}{5764--5773}.
\newblock
\href{https://doi.org/10.1109/ICCV.2019.00586}{doi:\nolinkurl{10.1109/ICCV.2019.00586}}


\bibitem[Wang et~al\mbox{.}(2024)]%
        {24aaaiskysrcipt}
\bibfield{author}{\bibinfo{person}{Zhecheng Wang}, \bibinfo{person}{Rajanie
  Prabha}, \bibinfo{person}{Tianyuan Huang}, \bibinfo{person}{Jiajun Wu}, {and}
  \bibinfo{person}{Ram Rajagopal}.} \bibinfo{year}{2024}\natexlab{}.
\newblock \showarticletitle{{SkyScript}: A Large and Semantically Diverse
  Vision-Language Dataset for Remote Sensing}.
\newblock \bibinfo{journal}{\emph{Proceedings of the AAAI Conference on
  Artificial Intelligence}} \bibinfo{volume}{38}, \bibinfo{number}{6}
  (\bibinfo{year}{2024}), \bibinfo{pages}{5805--5813}.
\newblock
\href{https://doi.org/10.1609/aaai.v38i6.28393}{doi:\nolinkurl{10.1609/aaai.v38i6.28393}}


\bibitem[Wei et~al\mbox{.}(2025)]%
        {wei2025hqclip}
\bibfield{author}{\bibinfo{person}{Zhixiang Wei}, \bibinfo{person}{Guangting
  Wang}, \bibinfo{person}{Xiaoxiao Ma}, \bibinfo{person}{Ke Mei},
  \bibinfo{person}{Huaian Chen}, \bibinfo{person}{Yi Jin}, {and}
  \bibinfo{person}{Fengyun Rao}.} \bibinfo{year}{2025}\natexlab{}.
\newblock \showarticletitle{{HQ-CLIP}: Leveraging Large Vision-Language Models
  to Create High-Quality Image-Text Datasets and {CLIP} Models}. In
  \bibinfo{booktitle}{\emph{Proceedings of the IEEE/CVF International
  Conference on Computer Vision (ICCV)}}. \bibinfo{pages}{22447--22456}.
\newblock
\href{https://doi.org/10.1109/ICCV51701.2025.02084}{doi:\nolinkurl{10.1109/ICCV51701.2025.02084}}


\bibitem[Yamaguchi et~al\mbox{.}(2025)]%
        {25cvprPostAlign}
\bibfield{author}{\bibinfo{person}{Shin'ya Yamaguchi}, \bibinfo{person}{Dewei
  Feng}, \bibinfo{person}{Sekitoshi Kanai}, \bibinfo{person}{Kazuki Adachi},
  {and} \bibinfo{person}{Daiki Chijiwa}.} \bibinfo{year}{2025}\natexlab{}.
\newblock \showarticletitle{Post-Pre-Training for Modality Alignment in
  Vision-Language Foundation Models}. In \bibinfo{booktitle}{\emph{Proceedings
  of the IEEE/CVF Conference on Computer Vision and Pattern Recognition
  (CVPR)}}. \bibinfo{pages}{4256--4266}.
\newblock
\href{https://doi.org/10.1109/CVPR52734.2025.00402}{doi:\nolinkurl{10.1109/CVPR52734.2025.00402}}


\bibitem[Yang et~al\mbox{.}(2024b)]%
        {24tgrsIERR}
\bibfield{author}{\bibinfo{person}{Lingling Yang}, \bibinfo{person}{Tongqing
  Zhou}, \bibinfo{person}{Wentao Ma}, \bibinfo{person}{Mengze Du},
  \bibinfo{person}{Lu Liu}, \bibinfo{person}{Feng Li}, \bibinfo{person}{Shan
  Zhao}, {and} \bibinfo{person}{Yuwei Wang}.} \bibinfo{year}{2024}\natexlab{b}.
\newblock \showarticletitle{Remote Sensing Image-Text Retrieval with
  Implicit-Explicit Relation Reasoning}.
\newblock \bibinfo{journal}{\emph{IEEE Transactions on Geoscience and Remote
  Sensing}}  \bibinfo{volume}{62} (\bibinfo{year}{2024}),
  \bibinfo{pages}{1--11}.
\newblock
\href{https://doi.org/10.1109/TGRS.2024.3466909}{doi:\nolinkurl{10.1109/TGRS.2024.3466909}}


\bibitem[Yang et~al\mbox{.}(2024a)]%
        {24tgrsMSCMAT}
\bibfield{author}{\bibinfo{person}{Rui Yang}, \bibinfo{person}{Shuang Wang},
  \bibinfo{person}{Yingping Han}, \bibinfo{person}{Yuanheng Li},
  \bibinfo{person}{Dong Zhao}, \bibinfo{person}{Dou Quan},
  \bibinfo{person}{Yanhe Guo}, \bibinfo{person}{Licheng Jiao}, {and}
  \bibinfo{person}{Zhi Yang}.} \bibinfo{year}{2024}\natexlab{a}.
\newblock \showarticletitle{Transcending Fusion: A Multiscale Alignment Method
  for Remote Sensing Image--Text Retrieval}.
\newblock \bibinfo{journal}{\emph{IEEE Transactions on Geoscience and Remote
  Sensing}}  \bibinfo{volume}{62} (\bibinfo{year}{2024}),
  \bibinfo{pages}{1--17}.
\newblock
\href{https://doi.org/10.1109/TGRS.2024.3496898}{doi:\nolinkurl{10.1109/TGRS.2024.3496898}}


\bibitem[Yang et~al\mbox{.}(2025)]%
        {25aclOMGM}
\bibfield{author}{\bibinfo{person}{Wei Yang}, \bibinfo{person}{Jingjing Fu},
  \bibinfo{person}{Rui Wang}, \bibinfo{person}{Jinyu Wang},
  \bibinfo{person}{Lei Song}, {and} \bibinfo{person}{Jiang Bian}.}
  \bibinfo{year}{2025}\natexlab{}.
\newblock \showarticletitle{{OMGM}: Orchestrate Multiple Granularities and
  Modalities for Efficient Multimodal Retrieval}. In
  \bibinfo{booktitle}{\emph{Proceedings of the 63rd Annual Meeting of the
  Association for Computational Linguistics (Volume 1: Long Papers)}}.
  \bibinfo{pages}{24545--24563}.
\newblock
\href{https://doi.org/10.18653/v1/2025.acl-long.1198}{doi:\nolinkurl{10.18653/v1/2025.acl-long.1198}}


\bibitem[Yuan et~al\mbox{.}(2023)]%
        {23tgrsPETL}
\bibfield{author}{\bibinfo{person}{Yuan Yuan}, \bibinfo{person}{Yang Zhan},
  {and} \bibinfo{person}{Zhitong Xiong}.} \bibinfo{year}{2023}\natexlab{}.
\newblock \showarticletitle{Parameter-Efficient Transfer Learning for Remote
  Sensing Image--Text Retrieval}.
\newblock \bibinfo{journal}{\emph{IEEE Transactions on Geoscience and Remote
  Sensing}}  \bibinfo{volume}{61} (\bibinfo{year}{2023}),
  \bibinfo{pages}{1--14}.
\newblock
\href{https://doi.org/10.1109/TGRS.2023.3308969}{doi:\nolinkurl{10.1109/TGRS.2023.3308969}}


\bibitem[Yuan et~al\mbox{.}(2022a)]%
        {22tgrsAMFMN}
\bibfield{author}{\bibinfo{person}{Zhiqiang Yuan}, \bibinfo{person}{Wenkai
  Zhang}, \bibinfo{person}{Kun Fu}, \bibinfo{person}{Xuan Li},
  \bibinfo{person}{Chubo Deng}, \bibinfo{person}{Hongqi Wang}, {and}
  \bibinfo{person}{Xian Sun}.} \bibinfo{year}{2022}\natexlab{a}.
\newblock \showarticletitle{Exploring a Fine-Grained Multiscale Method for
  Cross-Modal Remote Sensing Image Retrieval}.
\newblock \bibinfo{journal}{\emph{IEEE Transactions on Geoscience and Remote
  Sensing}}  \bibinfo{volume}{60} (\bibinfo{year}{2022}),
  \bibinfo{pages}{1--19}.
\newblock
\href{https://doi.org/10.1109/TGRS.2021.3078451}{doi:\nolinkurl{10.1109/TGRS.2021.3078451}}


\bibitem[Yuan et~al\mbox{.}(2022b)]%
        {21tgrsLW-MCR}
\bibfield{author}{\bibinfo{person}{Zhiqiang Yuan}, \bibinfo{person}{Wenkai
  Zhang}, \bibinfo{person}{Xuee Rong}, \bibinfo{person}{Xuan Li},
  \bibinfo{person}{Jialiang Chen}, \bibinfo{person}{Hongqi Wang},
  \bibinfo{person}{Kun Fu}, {and} \bibinfo{person}{Xian Sun}.}
  \bibinfo{year}{2022}\natexlab{b}.
\newblock \showarticletitle{A Lightweight Multi-Scale Crossmodal Text-Image
  Retrieval Method in Remote Sensing}.
\newblock \bibinfo{journal}{\emph{IEEE Transactions on Geoscience and Remote
  Sensing}}  \bibinfo{volume}{60} (\bibinfo{year}{2022}),
  \bibinfo{pages}{1--19}.
\newblock
\href{https://doi.org/10.1109/TGRS.2021.3124252}{doi:\nolinkurl{10.1109/TGRS.2021.3124252}}


\bibitem[Yuan et~al\mbox{.}(2022c)]%
        {22tgrsGaLR}
\bibfield{author}{\bibinfo{person}{Zhiqiang Yuan}, \bibinfo{person}{Wenkai
  Zhang}, \bibinfo{person}{Changyuan Tian}, \bibinfo{person}{Xuee Rong},
  \bibinfo{person}{Zhengyuan Zhang}, \bibinfo{person}{Hongqi Wang},
  \bibinfo{person}{Kun Fu}, {and} \bibinfo{person}{Xian Sun}.}
  \bibinfo{year}{2022}\natexlab{c}.
\newblock \showarticletitle{Remote Sensing Cross-Modal Text-Image Retrieval
  Based on Global and Local Information}.
\newblock \bibinfo{journal}{\emph{IEEE Transactions on Geoscience and Remote
  Sensing}}  \bibinfo{volume}{60} (\bibinfo{year}{2022}),
  \bibinfo{pages}{1--16}.
\newblock
\href{https://doi.org/10.1109/TGRS.2022.3163706}{doi:\nolinkurl{10.1109/TGRS.2022.3163706}}


\bibitem[Zhang et~al\mbox{.}(2026a)]%
        {26tommGNNRerank}
\bibfield{author}{\bibinfo{person}{Xuanmeng Zhang}, \bibinfo{person}{Minyue
  Jiang}, \bibinfo{person}{Zhedong Zheng}, \bibinfo{person}{Xiao Tan}, {and}
  \bibinfo{person}{Yi Yang}.} \bibinfo{year}{2026}\natexlab{a}.
\newblock \showarticletitle{Understanding Image Retrieval Re-Ranking: A Graph
  Neural Network Perspective}.
\newblock \bibinfo{journal}{\emph{ACM Transactions on Multimedia Computing,
  Communications, and Applications}} \bibinfo{volume}{22}, \bibinfo{number}{5},
  Article \bibinfo{articleno}{145} (\bibinfo{year}{2026}),
  \bibinfo{numpages}{19}~pages.
\newblock
\href{https://doi.org/10.1145/3803010}{doi:\nolinkurl{10.1145/3803010}}


\bibitem[Zhang et~al\mbox{.}(2023)]%
        {23RemoteSensingMTGFE}
\bibfield{author}{\bibinfo{person}{Xiong Zhang}, \bibinfo{person}{Weipeng Li},
  \bibinfo{person}{Xu Wang}, \bibinfo{person}{Luyao Wang},
  \bibinfo{person}{Fuzhong Zheng}, \bibinfo{person}{Long Wang}, {and}
  \bibinfo{person}{Haisu Zhang}.} \bibinfo{year}{2023}\natexlab{}.
\newblock \showarticletitle{A Fusion Encoder with Multi-Task Guidance for
  Cross-Modal Text--Image Retrieval in Remote Sensing}.
\newblock \bibinfo{journal}{\emph{Remote Sensing}} \bibinfo{volume}{15},
  \bibinfo{number}{18} (\bibinfo{year}{2023}), \bibinfo{pages}{4637}.
\newblock
\href{https://doi.org/10.3390/rs15184637}{doi:\nolinkurl{10.3390/rs15184637}}


\bibitem[Zhang et~al\mbox{.}(2026b)]%
        {26tipPerceiveIR}
\bibfield{author}{\bibinfo{person}{Xu Zhang}, \bibinfo{person}{Jiaqi Ma},
  \bibinfo{person}{Guoli Wang}, \bibinfo{person}{Qian Zhang},
  \bibinfo{person}{Huan Zhang}, {and} \bibinfo{person}{Lefei Zhang}.}
  \bibinfo{year}{2026}\natexlab{b}.
\newblock \showarticletitle{{Perceive-IR}: Learning to Perceive Degradation
  Better for All-in-One Image Restoration}.
\newblock \bibinfo{journal}{\emph{IEEE Transactions on Image Processing}}
  \bibinfo{volume}{35} (\bibinfo{year}{2026}), \bibinfo{pages}{2018--2033}.
\newblock
\href{https://doi.org/10.1109/TIP.2025.3566300}{doi:\nolinkurl{10.1109/TIP.2025.3566300}}


\bibitem[Zhang et~al\mbox{.}(2024)]%
        {24tgrsGeoRSCLIP}
\bibfield{author}{\bibinfo{person}{Zilun Zhang}, \bibinfo{person}{Tiancheng
  Zhao}, \bibinfo{person}{Yulong Guo}, {and} \bibinfo{person}{Jianwei Yin}.}
  \bibinfo{year}{2024}\natexlab{}.
\newblock \showarticletitle{{RS5M} and {GeoRSCLIP}: A Large-Scale
  Vision-Language Dataset and a Large Vision-Language Model for Remote
  Sensing}.
\newblock \bibinfo{journal}{\emph{IEEE Transactions on Geoscience and Remote
  Sensing}}  \bibinfo{volume}{62} (\bibinfo{year}{2024}),
  \bibinfo{pages}{1--23}.
\newblock
\href{https://doi.org/10.1109/TGRS.2024.3449154}{doi:\nolinkurl{10.1109/TGRS.2024.3449154}}


\bibitem[Zhao et~al\mbox{.}(2023)]%
        {23sigirMGRL}
\bibfield{author}{\bibinfo{person}{Shengwei Zhao}, \bibinfo{person}{Linhai Xu},
  \bibinfo{person}{Yuying Liu}, {and} \bibinfo{person}{Shaoyi Du}.}
  \bibinfo{year}{2023}\natexlab{}.
\newblock \showarticletitle{Multi-Grained Representation Learning for
  Cross-Modal Retrieval}. In \bibinfo{booktitle}{\emph{Proceedings of the 46th
  International ACM SIGIR Conference on Research and Development in Information
  Retrieval}}. \bibinfo{pages}{2194--2198}.
\newblock
\href{https://doi.org/10.1145/3539618.3592025}{doi:\nolinkurl{10.1145/3539618.3592025}}


\bibitem[Zheng et~al\mbox{.}(2025a)]%
        {25tmmCPPMN}
\bibfield{author}{\bibinfo{person}{Chengyu Zheng}, \bibinfo{person}{Xiu Li},
  \bibinfo{person}{Xinyue Liang}, \bibinfo{person}{Lei Huang},
  \bibinfo{person}{Shan Du}, \bibinfo{person}{Jie Nie}, {and}
  \bibinfo{person}{Junyu Dong}.} \bibinfo{year}{2025}\natexlab{a}.
\newblock \showarticletitle{Cross-Modal Progressive Perspective Matching
  Network for Remote Sensing Image-Text Retrieval}.
\newblock \bibinfo{journal}{\emph{IEEE Transactions on Multimedia}}
  \bibinfo{volume}{27} (\bibinfo{year}{2025}), \bibinfo{pages}{3966--3979}.
\newblock
\href{https://doi.org/10.1109/TMM.2025.3535365}{doi:\nolinkurl{10.1109/TMM.2025.3535365}}


\bibitem[Zheng et~al\mbox{.}(2025b)]%
        {25tgrsFSSN}
\bibfield{author}{\bibinfo{person}{Chengyu Zheng}, \bibinfo{person}{Jie Nie},
  \bibinfo{person}{Bo Yin}, \bibinfo{person}{Xiu Li}, \bibinfo{person}{Yuntao
  Qian}, {and} \bibinfo{person}{Zhiqiang Wei}.}
  \bibinfo{year}{2025}\natexlab{b}.
\newblock \showarticletitle{Frequency and Spatial-Domain Saliency Network for
  Remote Sensing Cross-Modal Retrieval}.
\newblock \bibinfo{journal}{\emph{IEEE Transactions on Geoscience and Remote
  Sensing}}  \bibinfo{volume}{63} (\bibinfo{year}{2025}),
  \bibinfo{pages}{1--13}.
\newblock
\href{https://doi.org/10.1109/TGRS.2025.3561626}{doi:\nolinkurl{10.1109/TGRS.2025.3561626}}


\bibitem[Zheng et~al\mbox{.}(2023a)]%
        {23tommSSJDN}
\bibfield{author}{\bibinfo{person}{Chengyu Zheng}, \bibinfo{person}{Ning Song},
  \bibinfo{person}{Ruoyu Zhang}, \bibinfo{person}{Lei Huang},
  \bibinfo{person}{Zhiqiang Wei}, {and} \bibinfo{person}{Jie Nie}.}
  \bibinfo{year}{2023}\natexlab{a}.
\newblock \showarticletitle{Scale-Semantic Joint Decoupling Network for
  Image-Text Retrieval in Remote Sensing}.
\newblock \bibinfo{journal}{\emph{ACM Transactions on Multimedia Computing,
  Communications, and Applications}} \bibinfo{volume}{20}, \bibinfo{number}{1},
  Article \bibinfo{articleno}{4} (\bibinfo{year}{2023}),
  \bibinfo{numpages}{20}~pages.
\newblock
\href{https://doi.org/10.1145/3603628}{doi:\nolinkurl{10.1145/3603628}}


\bibitem[Zheng et~al\mbox{.}(2023b)]%
        {23sensorsFAAMI}
\bibfield{author}{\bibinfo{person}{Fuzhong Zheng}, \bibinfo{person}{Xu Wang},
  \bibinfo{person}{Luyao Wang}, \bibinfo{person}{Xiong Zhang},
  \bibinfo{person}{Hongze Zhu}, \bibinfo{person}{Long Wang}, {and}
  \bibinfo{person}{Haisu Zhang}.} \bibinfo{year}{2023}\natexlab{b}.
\newblock \showarticletitle{A Fine-Grained Semantic Alignment Method Specific
  to Aggregate Multi-Scale Information for Cross-Modal Remote Sensing Image
  Retrieval}.
\newblock \bibinfo{journal}{\emph{Sensors}} \bibinfo{volume}{23},
  \bibinfo{number}{20} (\bibinfo{year}{2023}), \bibinfo{pages}{8437}.
\newblock
\href{https://doi.org/10.3390/s23208437}{doi:\nolinkurl{10.3390/s23208437}}


\bibitem[Zhou et~al\mbox{.}(2025b)]%
        {25aclMMRet}
\bibfield{author}{\bibinfo{person}{Junjie Zhou}, \bibinfo{person}{Yongping
  Xiong}, \bibinfo{person}{Zheng Liu}, \bibinfo{person}{Ze Liu},
  \bibinfo{person}{Shitao Xiao}, \bibinfo{person}{Yueze Wang},
  \bibinfo{person}{Bo Zhao}, \bibinfo{person}{Chen~Jason Zhang}, {and}
  \bibinfo{person}{Defu Lian}.} \bibinfo{year}{2025}\natexlab{b}.
\newblock \showarticletitle{MegaPairs: Massive Data Synthesis for Universal
  Multimodal Retrieval}. In \bibinfo{booktitle}{\emph{Proceedings of the 63rd
  Annual Meeting of the Association for Computational Linguistics (Volume 1:
  Long Papers)}}. \bibinfo{publisher}{Association for Computational
  Linguistics}, \bibinfo{address}{Vienna, Austria},
  \bibinfo{pages}{19076--19095}.
\newblock
\href{https://doi.org/10.18653/v1/2025.acl-long.935}{doi:\nolinkurl{10.18653/v1/2025.acl-long.935}}


\bibitem[Zhou et~al\mbox{.}(2025a)]%
        {25grslCFITR}
\bibfield{author}{\bibinfo{person}{Wenqian Zhou}, \bibinfo{person}{Hanlin Wu},
  {and} \bibinfo{person}{Pei Deng}.} \bibinfo{year}{2025}\natexlab{a}.
\newblock \showarticletitle{Toward Efficient and Accurate Remote Sensing
  Image--Text Retrieval with a Coarse-to-Fine Approach}.
\newblock \bibinfo{journal}{\emph{IEEE Geoscience and Remote Sensing Letters}}
  \bibinfo{volume}{22} (\bibinfo{year}{2025}), \bibinfo{pages}{1--5}.
\newblock
\href{https://doi.org/10.1109/LGRS.2024.3494543}{doi:\nolinkurl{10.1109/LGRS.2024.3494543}}


\bibitem[Zhu et~al\mbox{.}(2025)]%
        {25nnUSSL}
\bibfield{author}{\bibinfo{person}{Jie Zhu}, \bibinfo{person}{Jianan Liu},
  \bibinfo{person}{Shufang Wu}, {and} \bibinfo{person}{Feng Zhang}.}
  \bibinfo{year}{2025}\natexlab{}.
\newblock \showarticletitle{Unified Semantic Space Learning for Cross-Modal
  Retrieval}.
\newblock \bibinfo{journal}{\emph{Neural Networks}}  \bibinfo{volume}{191}
  (\bibinfo{year}{2025}), \bibinfo{pages}{107756}.
\newblock
\href{https://doi.org/10.1016/j.neunet.2025.107756}{doi:\nolinkurl{10.1016/j.neunet.2025.107756}}


\end{thebibliography}

\end{document}